\pdfoutput=1
\documentclass[11pt]{article}

\usepackage[dvipsnames]{xcolor}
\newcommand{\rev}[1]{\textcolor{black}{#1}}
\usepackage[final]{acl}

\usepackage{times}
\usepackage{latexsym}

\usepackage{enumitem}

\usepackage{tabularx}
\usepackage{multirow}
\usepackage{booktabs}
\usepackage{graphicx}
\usepackage{amsmath,amssymb,amsfonts}
\usepackage{url}
\usepackage{subfigure}
\usepackage[breakable]{tcolorbox}
\definecolor{titleblockcolor}{HTML}{353535}
\definecolor{textblockcolor}{HTML}{FFFFFF}
\newenvironment{block}[2][]{
  \begin{tcolorbox}[adjusted title=#2, fonttitle={\small\bfseries}, colback={textblockcolor}, colframe={titleblockcolor}, coltitle={white}, arc=0pt,
  outer arc=0pt, left=1pt, right=1pt, fontupper=\small, #1, breakable]
}{\end{tcolorbox}}
\newtcbox{\highlight}[1][red]
  {on line, arc = 0pt, outer arc = 0pt,
    colback = #1!10!white, colframe = #1!50!black,
    boxsep = 0pt, left = 1pt, right = 1pt, top = 2pt, bottom = 2pt,
    boxrule = 0pt, bottomrule = 1pt, toprule = 1pt}

\usepackage[T1]{fontenc}
\usepackage[utf8]{inputenc}

\usepackage{microtype}

\usepackage{inconsolata}

\usepackage{graphicx}

\title{KERAG: Knowledge-Enhanced Retrieval-Augmented Generation for Advanced Question Answering}

\author{Yushi Sun$^{1}$, Kai Sun$^{2}$, Yifan Ethan Xu$^{2}$, Xiao Yang$^{2}$, Xin Luna Dong$^{2}$, Nan Tang$^{3}$\thanks{Corresponding author.}, Lei Chen$^{3,1}$ \\
$^{1}$HKUST, $^{2}$Meta Reality Labs, $^{3}$HKUST(GZ)\\
$^{1}$\texttt{\{ysunbp, leichen\}@cse.ust.hk}, 
$^{2}$\texttt{\{sunkaicn, ethanxu, xiaoyangfb, lunadong\}@meta.com},\\
$^{3}$\texttt{\{nantang, leichen\}@hkust-gz.edu.cn}}

\begin{document}
\maketitle
\begin{abstract}
Retrieval-Augmented Generation (RAG) mitigates hallucination in Large Language Models (LLMs) by incorporating external data, with Knowledge Graphs (KGs) offering crucial information for question answering. Traditional Knowledge Graph Question Answering (KGQA) methods rely on semantic parsing, which typically retrieves knowledge strictly necessary for answer generation, thus often suffer from low coverage due to rigid schema requirements and semantic ambiguity. We present {\sc KERAG}\footnote{All the code and data were done by HKUST: \url{https://github.com/ysunbp/KERAG}.}, a novel KG-based RAG pipeline that enhances QA coverage by retrieving a broader subgraph likely to contain relevant information. Our retrieval-filtering-summarization approach, combined with fine-tuned LLMs for Chain-of-Thought reasoning on knowledge sub-graphs, reduces noises and improves QA for both simple and complex questions. Experiments demonstrate that KERAG surpasses state-of-the-art solutions by about 7\% in quality and exceeds GPT-4o (Tool) by 10-21\%. 
\end{abstract}

%!TEX root = ../acl_latex.tex
\section{Introduction}
\label{sec:intro}

Retrieval-Augmented Generation (RAG) has been extensively explored recently in both academia and industry, as a means to mitigate the hallucination issues of Large Language Models (LLMs)~\cite{sun2024large, yang2024crag, fan2024survey, ni2025towards, sun2025knowledge}.
RAG enhances LLM outputs by incorporating relevant data from external sources, thereby grounding the generated content with factual knowledge. Knowledge graphs serve as a crucial data source, providing efficient access to rich and precise information for question answering~\cite{Dong2024QA, hogan2025large}.

\begin{figure}[t!]
  \centering

\includegraphics[width=1.0\linewidth]{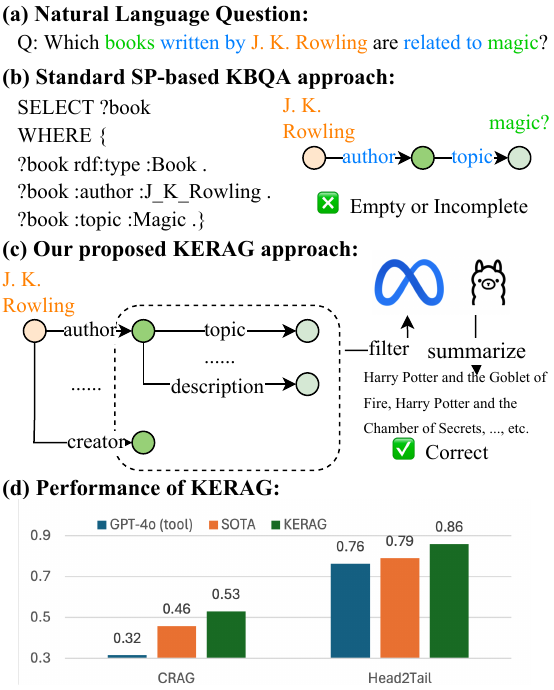}
  \caption{(a) An exemplar query. (b) On the left, we show the SPARQL query generated by the triple-level SP-based approach. On the right, we display the linked KG content. This design leads to empty or incomplete answers since: 1) The book entities do not necessarily have an attribute named ``topic''; 2) ``magic'' may not be explicitly presented in the topic attributes, but may be embedded in other attributes such as description. (c) The rationale behind our KERAG approach: entity-level SP with LLM-based post-processing of the retrieved KG content. (d) Truthfulness of KERAG against state-of-the-arts (truthfulness=accuracy-hallucination).}

  \label{fig:motivation}
\end{figure}

Knowledge Graph Question Answering (KGQA) is a well-established field that has been extensively studied long before the advent of LLMs. Most solutions utilize the technique of {\em semantic parsing}, where a natural language question is converted into a structured query, in the form of either a SPARQL query~\cite{yih2016value, das2021case, gu2022arcaneqa, shu2022tiara, hu2022logical, xie2022unifiedskg,  xu2023fine, zhang2023fc}, or a retrieval path that first identifies the queried entity and then follows a particular predicate to a neighboring entity or attribute value as the answer~\cite{yih2015semantic, luo2018knowledge, lanknowledge,yu2022decaf}. When successful, these solutions produce precise and concise answers. However, they require rigorous schemas (ontologies) and have low tolerance to parsing errors (see Figure~\ref{fig:motivation} for an example). As a result, whereas KGQA generally provides highly accurate answers, it often suffers from low coverage, missing answers for many questions. \rev{With LLMs emerging, recent works~\cite{sunthink, ma2025thinkongraph} leverage the reasoning ability of LLMs to generate {\em multiple} retrieval paths; however, the small number of paths (e.g., up to 3 in~\cite{ma2025thinkongraph}) does not fully address the retrieval recall issue.}

LLMs, with their strong summarization and reasoning capabilities, offer a new perspective to pursue KGQA \rev{and allow us to push {\em reasoning} from retrieval-path generation to the summarization step. At retrieval time, instead of following up to a few retrieval {\em paths}, we gather from the neighborhood of the topic entity \textit{all} information potentially relevant to the question. At answer generation time, we leverage LLM's reasoning capabilities to identify relevant information and generate accurate answers. By focusing on the whole neighborhood rather than a few KG paths,} the complexity of generating logical forms for KG querying is significantly reduced and the retrieval recall can significantly increase. 

%LLMs, with their strong summarization and reasoning capabilities, offer a new perspective to pursue KGQA. \rev{Several recent works~\cite{sunthink, ma2025thinkongraph}, proposed to leverage the reasoning ability of LLMs to autonomously obtain relevant paths from the KGs for QA. Despite the state-of-the-art performance in several benchmarks, we argue that using LLMs to extract a small number of paths from KG as relevant information for QA potentially limits the retrieval recall and thus affecting the accuracy of KGQA. Therefore,} instead of adhering to a strict querying method, we relax the retrieval process to gather all relevant information about the \rev{topic entity} mentioned in the question, thereby lifting recall. We then leverage the reasoning capabilities of LLMs to \rev{filter the retrieved content and} generate accurate answers. \rev{By focusing on entity-level neighborhood rather than KG path retrieval,} the complexity of generating logical forms for KG querying is significantly reduced and the retrieval recall can significantly increase. 

Although the idea appears straightforward, there are three prominent challenges. First, entities, particularly head entities, can be associated with a substantial amount of knowledge (2 million triples per head entity~\cite{dong2014data}). All knowledge may not fit within the context window, and even if it does, the noises can lead to confusion during summarization. Second, some questions require information from multi-hop neighbors, complicating the determination of retrieval boundaries, and further increasing the volume of retrieved information. Lastly, complex questions that necessitate aggregation and reasoning remain challenging.
 
We proposed KERAG, a novel Knowledge-Enhanced RAG pipeline that employs a {\em retrieval-filter-summarization} paradigm. The \textit{retrieval} part gathers information about the \rev{topic entity} mentioned in the question; the \textit{filtering} part eliminates information irrelevant to the question; and the \textit{summarization} part reasons over the remaining information to generate the answer. Rather than retrieving the entities and triples strictly necessary for generating an answer, our method retrieves a broader subgraph that is likely to contain relevant information for answer generation.

There are two key innovations that ensure high-quality answers. First, in our planning phase we interleave multi-hop neighborhood retrieval with filtering according to the KG schema, allowing us to filter at the predicate level rather than overwhelming the summarizer with irrelevant triples, and meanwhile ensuring that all information necessary for question answering is retrieved from the neighborhood. Second, we fine-tune the LLM for Chain-of-Thought (CoT) reasoning over the neighborhood sub-graph, enabling it to generate accurate answers for both simple questions and complex questions. An additional advantage of our method is its applicability not only to graph databases that support SPARQL, but also to API-based knowledge-providing systems, which are commonly used in production~\cite{yang2024crag}. 

To summarize, we make three contributions.
\begin{itemize}[itemsep=0pt, topsep=0pt]
    \item We introduce a novel KG-based RAG pipeline, KERAG, which retrieves information at the entity level, rather than at the triple level as in traditional semantic parsing solutions. This approach significantly enhances QA coverage. 
    \item We design a retrieval-filter-summarization \rev{paradigm}, complemented by fine-tuning LLMs for Chain-of-Thought-based summarization. This solution reduces the noises in summarization and improves answer generation for both simple and complex questions.
    \item We conducted extensive experiments on two KG benchmarks (API-based and SPARQL-based), showing that KERAG outperforms state-of-the-art solutions by ~7\% in quality, and outperforms GPT-4o (Tool) by 10-21\%.
\end{itemize}

%!TEX root = ../acl_latex.tex
\section{Related Work}
One major line of research of KGQA is semantic-parsing-based (SP-based) KGQA approaches, which primarily transform natural language queries into logical forms and initiate structured queries to access the knowledge graph content. The SP-based KGQA approaches can be further divided into multi-step and seq2seq approaches. The multi-step approaches generally formulate the semantic parsing problem as a multi-step search problem: identifying the core entity, generating and expanding the query graph based on the relationships between the attributes of the entities and the predicates of the queries~\cite{lanknowledge, lan2020query, oguz2022unik}. The seq2seq approaches primarily consider the semantic parsing problem in a seq2seq manner with the help of fine-tuned language models: directly generating the complete semantic expression based on the query and retrieving the relevant knowledge graph content~\cite{das2021case,  shu2022tiara, hu2022logical, xie2022unifiedskg, yu2022decaf, zhang2023fc, gu2023don, xu2023fine}. Specifically, %~\cite{das2021case} introduces a reasoning framework based on cases that first identifies queries and logical forms with semantically similar components. By retrieving analogous instances from maintained memory, they combine these components to generate new logical forms for complex questions.
%~\cite{ye2022rng} combines generation and ranking: the authors propose generating a set of candidate logical expressions and then ranking them to select the best logical form.~\cite{cao2022program} proposes first transforming the query into a high-level sketch and then refining the detailed arguments in the sketch by parsing the query with a specific argument parser.
%TIARA~\cite{shu2022tiara} offers a novel technique to enhance KGQA by retrieving the most relevant KG context from multiple granularities such as entities, logical expressions, and schema elements. 
Pangu~\cite{gu2023don} presents an innovative framework for grounded language understanding that integrates a symbolic agent with a neural language model. This system enables the stepwise formulation of valid plans while employing the language model to evaluate the viability of these logical plans in an online manner. 
WikiSP is a novel few-shot seq2seq semantic parser developed by~\cite{xu2023fine}, which fine-tunes a Llama model on modified SPARQL queries.

% IR-based approach
Another line of research in KGQA is \rev{the information-retrieval-based KGQA approaches}, where a subgraph is generated based on the entities and relations in the query, graph matching and retrieval techniques are applied to retrieve relevant graphs from the KG to answer the query~\cite{saxena2020improving, he2021improving, sen2021expanding, shi2021transfernet, mavromatis2022rearev, zhang2022subgraph}. For instance, %KV-Mem~\cite{miller2016key} employs a memory architecture based on key-value pairs to improve document understanding and facilitate question-answering. 
%PullNet~\cite{sun2019pullnet} introduces an approach that progressively builds a question-oriented subgraph by utilizing knowledge graphs and textual information. %EmbedKGQA~\cite{saxena2020improving} presents a technique that leverages knowledge graph embeddings to tackle issues related to the sparsity of knowledge graphs.
%NSM+h~\cite{he2021improving} presents a novel teacher-student architecture aimed at enhancing multi-hop knowledge-based question answering. In this framework, the teacher network generates intermediate supervisory signals by employing both forward and backward reasoning techniques, which subsequently improves the reasoning proficiency of the student network. TransferNet~\cite{shi2021transfernet} presents a model that integrates a clear, attention-driven methodology with the capability to manage both label and textual relationships. 
\cite{zhang2022subgraph} introduces a novel approach with a trainable subgraph retriever (SR) designed to operate separately from the reasoning process, identifying relevant subgraphs and prioritizing smaller and more relevant subgraphs. The subgraphs identified by our approach are more relaxed compared to~\cite{zhang2022subgraph}, which provide a broader context for the summarizer to generate accurate answers.% Besides, as suggested by~\cite{xu2023fine}, the IR-based approaches cannot handle entire categories of questions and have poor interpretability. 

% LLM-based approach
Thanks to the strong reasoning ability of language models, an increasing number of works start to seek or analyze the assistance of language models~\cite{sun2023reca, sun2024large, li2022fake, li2024cllmate,li2025pipe,wang2024scidasynth, chen2025unmasking}.
The KGQA community also focuses on LLM-based approaches recently, which utilize large language models (LLMs) to perform KGQA~\cite{sunthink, ma2025thinkongraph, jiang2023structgpt}. As for the LLM-based KGQA approaches, several state-of-the-art LLMs such as Llama-3.1s~\cite{dubey2024llama} support tool calling, which provides the opportunity to directly prompt the LLMs to generate API calls or SPARQL queries. However, retrieving directly with the tool call plan given by the LLM would propagate planning errors into the retrieval process, resulting in insufficient KG content being retrieved. We provide a comparison of performance between our KERAG pipeline and the Llama-3.1-70B-Instruct (tool) model (which can freely plan and initiate multiple rounds of API calls) in Tables~\ref{tab:baseline_crag} and~\ref{tab:baseline_htt}. 

Beyond naive LLM tool calling, ToG~\cite{sunthink} introduces beam search on the knowledge graph to iteratively explore reasoning paths for LLMs. ToG-2~\cite{ma2025thinkongraph} links to external Wikipedia pages to get additional context, the retrieval and reasoning are conducted between KG and external pages. 
%Pangu~\cite{gu2023don} presents an innovative framework for grounded language understanding that integrates a symbolic agent with a neural language model. This system enables the stepwise formulation of valid plans while employing the language model to evaluate the viability of these plans in an online manner. 
StructGPT~\cite{jiang2023structgpt} designs an interface for data retrieval and a unique process that involves invoking, linearizing, and generating outputs. 
\rev{Our method is fundamentally different: these approaches retrieve information from KGs based on retrieval paths, whereas KERAG retrieves all potentially relevant information in a neighborhood, thus improving on the retrieval recall (\textasciitilde11\% over ToG, Section~\ref{sec:exp-results}). As such, KERAG achieves a higher accuracy on all benchmarks, improving over ToG and ToG-2 by up to 20\% and 4\% respectively (Table~\ref{tab:qald-results}), and outperforming StructGPT by 7\% in truthfulness (Table~\ref{tab:baseline_htt}). Additionally, KERAG has the flexibility to apply both on KGs and APIs, allowing experiments on datasets like Head2Tail~\cite{sun2024head} and CRAG~\cite{yang2024crag}. We make detailed comparisons in Appendix~\ref{appendix:compare}.}
\section{Problem Statement}
\label{sec:problem}
% 1. I think our problem definition should assume a typical KG, instead of APIs. Otherwise it causes confusion.
% 2. Before Sec 2 I think we shall have related work, since this is a common topic.
% 3. Sec 2. Please remove current definitions but just define WHAT is KG-based RAG. Don't put any "HOW" in your problem defn.
% 4. Sec 3. 1) high-level architecture needs to be more formal; 2) each subsection shall start w. input/output (defn of the sub-problem), and then define the methods formally; 3) We shall highlight what is innovative in each subsection; 4) we shall repeat the challenges and highlight our solutions 
% 5. For each table/figure, give exp result summaries
% 6. Do we have baselines for SPARQL?

%\nan{
%First, define KB.
%Second, define QA with KBs.
%Then, say there are two typical ways of retrieving information from KBs: one is through SPARQL queries, and the other is through APIs.
%We can add a discussion that APIs are widely used in industries to access KBs, and give several citations or links.
%}

\noindent \textbf{Knowledge Graph:} 
A {\em Knowledge Graph (KG)} $K$ can be defined as $K=(E,R,D)$, where $E$ is a set of \textit{entities}, $R$ is a set of \textit{relations} and \rev{optionally} $D$ is a set of \textit{domains} covered by $K$. Each entity $e_i \in E$ can have properties. Each relation $r_j \in R$ can be represented as a binary relation $R:E \times E \rightarrow \mathbb{B}$. A KG is normally accompanied with an {\em ontology}, describing the entity types and relationships between different types of entities. %A query interface takes user queries as input and returns the results from KG, where a query is expressed in a formal language such as SPARQL, and the results may include entities, relationships, or inferred knowledge.

\smallskip
\noindent \textbf{KG-based RAG:} Given a natural language question $Q$ with a KG $K$, \textit{KG-based RAG} aims to answer the question $Q$ leveraging knowledge in $K$.

%!TEX root = ../acl_latex.tex
\section{Methodology}

\begin{figure*}[t!]
  \centering

\includegraphics[width=0.75\linewidth]{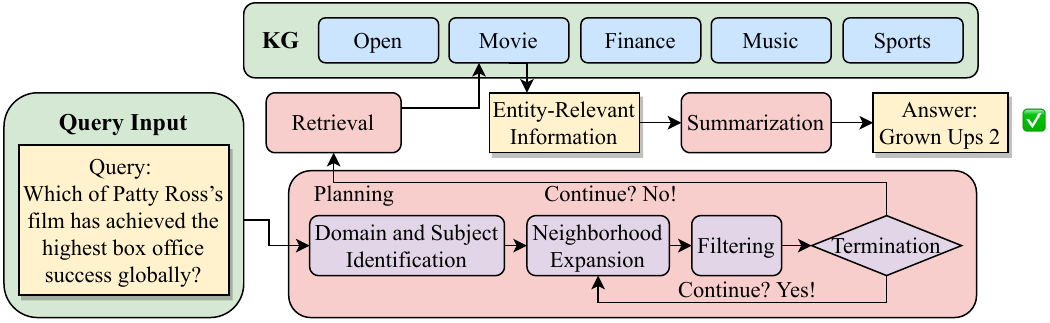}
  \caption{The overall pipeline of our KERAG approach.}
  \label{fig:pipe}

\end{figure*}

%\nan{We need to first give a high level idea of our approach. }

%\nan{This is an overview. Need to revise to better align with the figure.}
\subsection{Solution overview}
We start with an overview of our solution. Given a question $Q$ and a knowledge graph $K$, KERAG retrieves knowledge from $K$ and uses it to generate the answer for $Q$. As illustrated in Figure~\ref{fig:pipe}, the pipeline contains three major steps.
\begin{enumerate}[itemsep=0pt, topsep=0pt]
    \item {\bf Planning:} The \textit{Planning} step formulates a retrieval plan by selecting a \rev{topic entity} $E$ as a starting point and determining the scope of its neighborhood for retrieval. The scope includes the number of hops $h$ for neighborhood exploration and relations $\bar R$ to filter.
    %the domain $D$ of the question, 
    %\item {\bf Intent identification:} We first conduct entity-level parsing on the question $Q$, to decide the {\em topic entity} $E$ mentioned in $Q$ and the {\em domain} $D$, where the domain shall be consistent with the KG schema. 
    %\item {\bf Knowledge retrieval:} We query the knowledge graph $K$, iteratively retrieve $E$'s neighborhood knowledge until reaching the queried domain $D$.
    %\item {\bf Predicate filtering:} For each step of retrieval, we filter at the relation level for knowledge irrelevant to the question.
    \item {\bf Retrieval:} The \textit{Retrieval} step translates the retrieval plan to KG queries or API calls to retrieve relevant knowledge.
    \item {\bf Summarization:} Finally, the {\em Summarization} step reasons over the retrieved knowledge and produces the final answer.  
\end{enumerate}

The Retrieval step is fairly easy and varies based on the underlying KG system. In the rest of the section, we describe the planning and summarization steps in detail, with a focus on how they address the three main challenges: managing {\em knowledge overloading}, determining {\em multi-hop retrieval boundaries}, and handling {\em complex questions}. 

%\nan{The biggest problem of the following four subsections is that we did not describe our contributions.}

\subsection{Planning} %Intent identification} %Parser}

% Here, maybe need some concrete examples to show why domain and time is important.

The planning step takes a question $Q$ as input, and decides a retrieval plan as a quadruplet $(D, E, \bar R, h)$, where $D$ denotes the domain, $E$ denotes the topic entity as the starting point, $\bar R$ denotes the relations to filter, and $h$ denotes the number of hops to explore in the neighborhood.

We distinguish the retrieval plan from a retrieval query (e.g., SPARQL) or a retrieval path generated by traditional semantic parsing. Normal retrieval queries return a particular node (or a set of nodes) strictly necessary to answer the question. Our retrieval plan, instead, returns a neighborhood scope that might be needed for answering a question; only relations that are absolutely irrelevant to a question are excluded (e.g., \textit{Oscar awards} for a question regarding movie \textit{box office}). 

\begin{figure}[t!]
  \centering
\includegraphics[width=1.0\linewidth]{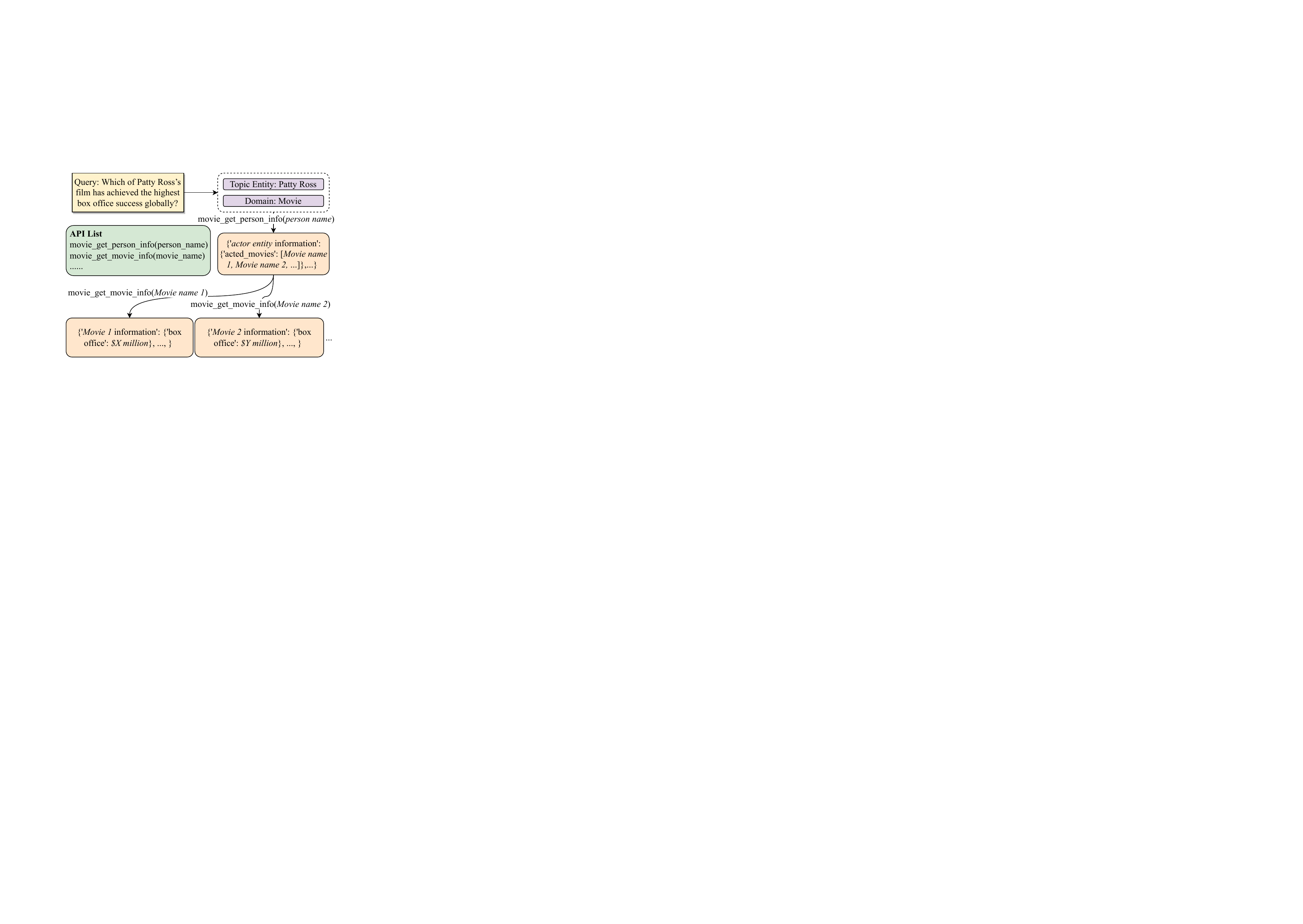}
  \caption{\rev{An exemplar query illustrating the planning.}}
  \label{fig:retrieve}

\end{figure}

Our planning proceeds in four steps.
\begin{itemize}[itemsep=0pt, topsep=0pt]
    \item {\em Step 1:} We start with prompt to decide the topic entity $E$ and the domain $D$ from the input question $Q$. As shown in Figure~\ref{fig:retrieve}, {\em "Which of Patty Ross’s film has achieved the highest box office success globally?"}. The topic entity is {\em Patty Ross} and the queried domain is {\em Movie}. 
    \item {\em Step 2:} Knowing the type of the topic entity, we leverage the schema to explore one hop in the neighborhood, identifying the relations and the types of the neighbors. From our example entity {\em Patty Ross}, the neighborhood includes relations such as acted movies, birthday, etc., with neighbors of types \textit{entity (movie)}, \textit{attribute (date)}, etc.
    \item {\em Step 3:} We feed the neighborhood information into an LLM to decide which relations are irrelevant, and whether enough information has been collected to answer question $Q$. For example, facts regarding {\em birth\_date} and {\em birth\_place} are irrelevant to answer the question; just knowing the movie neighbors without further information about these movies is not enough to answer the question. 
    \item {\em Step 4:} We terminate if LLM decides that there is enough information to answer the question, updating $h$ (\#hops) accordingly; otherwise, we go to Step 2 to explore the next hop of the neighborhood, unless we have exhausted relevant relations or reached a predetermined maximum number of hops. In our example, we will continue to retrieve information about movies, including their box office, necessary to answer the question. 
\end{itemize}

Next, we describe each step in more detail. 

\paragraph{Domain and Subject Identification:} 
%We start with deciding the main topic entity $E$ and the domain $D$. %Take as an example the query {\em "how many points did bulls put up in their game?"} The topic entity is {\em Chicago Bulls} and the queried domain is {\em Basketball}. 
%Correctness of this step sets the ceiling of overall accuracy. This is because 1) a misunderstood subject entity will lead to wrong retrieval results, and such errors are hard to recover in later steps; 2) some KG provides different APIs for different domains (\eg CRAG), thus a misunderstood domain can even lead to empty retrieval results. 
We leverage LLMs for this step. 
%Despite LLM's supreme understanding capability, this task is not necessarily trivial. In the given example, the entity name is abbreviated in the question, and domain knowledge is needed to realize that the domain is basketball. 
We apply few-shot learning and include necessary schema information to guide LLM, including the explanation and examples of domains. We use the following template (Planning Prompts Skeleton in Appendix~\ref{appendix:template}, details in the Appendix~\ref{appendix:prompt-parser}).

\begin{comment}
``\textit{You are given a Query [$Q_i$], please extract the main entity [$e_i$] from the Query.}''

``\textit{Determine the domain the query is about. The domain should be one of the following: [Set of KG domains].}''
\end{comment}

\paragraph{Neighborhood Expansion:} This step simply explores the schema of the knowledge graph, to understand relations and entity types in the next hop of the neighborhood to prepare the next step. %the purpose of retrieving neighborhood content (\textbf{Retrieval}).

\paragraph{Filtering:} This step decides 1) which relations in the obtained neighborhood content are absolutely irrelevant to answer the question $Q$ thus can be filtered, and 2) whether we have obtained enough information for question answering, addressing both the {\em knowledge overloading} challenge and {\em multi-hop boundary} challenge. We prepare both LLM-based filtering and similarity-based filtering~\cite{karpukhin2020dense} to filter according to the semantics of the question and the ontology. We apply the following LLM-based filtering template (Filtering Prompts Skeleton in Appendix~\ref{appendix:template}, details in the Appendix~\ref{appendix:prompt-filterer}):

\paragraph{Termination:} This step takes the LLM output to decide whether to continue planning or terminate. Once the LLM decides all necessary information is collected for question answering, we can proceed to the retrieval step. % it feeds the retrieved information into a prompt for answer generation (Detailed in Appendix~\ref{appendix:prompt-terminate}).

\smallskip
Our planning step has the following advantages. First, we enlarge the scope of retrieval compared to traditional semantic parsing, thus allowing us to lift the retrieval recall. Second, we allow multi-hop retrieval, such that we are able to answer complex questions regarding multi-hop reasoning. Third, we apply filtering to remove irrelevant content to avoid distracting the final summarization step. Finally, hopping and filtering decisions are made according to the ontology, significantly reduced the cost for full data retrieval (in case ontology is absent, we can easily adjust the algorithm to retrieve and reason about the data in each hop of neighborhood). On the {\sc CRAG} dataset, multi-hop retrieval improved truthfulness by 5.5\% and filtering improved by 3.9\% (Table~\ref{tab:ablation_crag}).

%\yushi{Modify the writing, we add a module to determine the multi-hop boundary.}

%The key reason we do not rely on the tool-calling capabilities provided by the LLM to directly retrieve KG content is due to error propagation considerations: retrieving directly with the tool call plan given by the LLM would propagate planning errors into the retrieval process, resulting in insufficient KG content being retrieved to answer the query. We provide a comparison of performance between our KERAG pipeline and the Llama-3.1-70B-Instruct (tool) model (which can freely plan and initiate multiple rounds of API calls) in Table~\ref{tab:baseline_benchmarking_crag} to illustrate this consideration.

%\noindent \textbf{Content-based filtering: }The content-based filtering is applicable to general interfaces, such as the SPARQL-based query interface. Specifically, we use DPR~\cite{karpukhin2020dense} to compute and rank the similarity between the candidate predicates in the retrieved triples and the query with the main entity removed. Top-K triples are preserved as the refined KG content.

%In summary, two different filtering techniques are introduced to handle cases where overwhelming KG content is retrieved. Our filterer design controls the knowledge volume present to the summarizer, which resolves the two challenges associated with the head entities and multi-hop retrieval. 

\subsection{CoT Summarization}
The final {\em Summarization} step takes the question $Q$ and the retrieval results, % $\bar{C_i}$, 
applies CoT reasoning to generate the answer. The key challenge is to answer complex questions leveraging the retrieved knowledge, oftentimes still in large volume. 

Complex questions, such as aggregation questions (e.g., \textit{What is the total points scored by the New Orleans Pelicans in 2022-01?}) and reasoning questions (e.g., \textit{Is memory management in Python handled through garbage collection?}), often require cross-referencing multiple pieces of KG information. We employ CoT prompting for the reasoning, the template is as follows (Summarization Prompts Skeleton in Appendix~\ref{appendix:template}, details in the Appendix~\ref{appendix:prompt-summarizer}). 

\begin{figure}[t!]
  \centering
  
\includegraphics[width=0.9\linewidth]{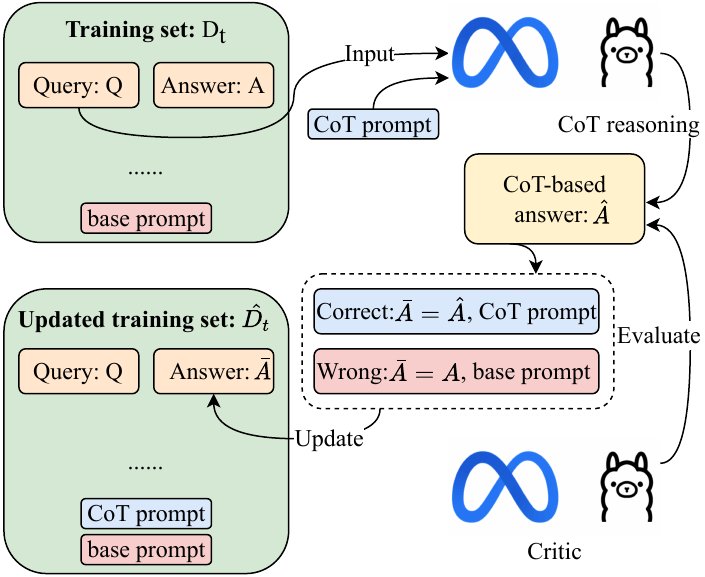}
  \caption{Generation of the fine-tuning data.}
  \label{fig:summarize}

\end{figure}

To further enhance performance, we fine-tune the CoT-based summarizer. To avoid the difficulties of manually labeling data~\cite{sun2024cross}, we proposed the following method to generate the fine-tuning data and present the details in Figure~\ref{fig:summarize}. 
\begin{enumerate}[itemsep=0pt, topsep=0pt]
    \item For each query $Q$ in the training set, we first prompt an LLM model to generate CoT-style reasoning together with the answer $\hat{A}$.
    \item We then prompt LLM to decide the correctness of $\hat{A}$ by comparing with the ground truth answer $A$.
    \item Based on the correctness judgment, we enhance the training data as follows. If the generated answer $\hat{A}$ is correct, we will take $\hat{A}$ together with the generated reasoning as the ground truth; otherwise, we retain the original ground truth and replace the CoT prompt with stand non-CoT prompt.
    \item The queries, the updated prompts, and the updated answers formulate our updated training set for Supervised Fine Tuning (SFT).
\end{enumerate}  

Our proposed CoT-based summarizer together with the fine-tuning data generation scheme is the biggest contributor to the end-to-end answer quality improvement. On the {\sc CRAG} dataset, using plain CoT reasoning refrained from generating hallucinated answers for 18\% of questions (mostly saying "I don't know" instead), and our fine-tuning further increased accuracy by 10\% (Table~\ref{tab:ablation_crag}).

%\subsection{Generalizability}
%\label{sec:generalize}

%In this section, we discuss how to generalize our method from API-based KGs to SPARQL-based KGs. The first change lies in the retrieval component. Unlike the relatively strict schema requirements imposed by the API functions, the SPARQL language is more flexible. This flexibility makes it difficult to effectively terminate the retrieval process using the schema's own requirements for parameters. Therefore, we limit the maximum number of hops as an alternative termination condition for SPARQL-based KGs. As for the filtering component, the tool-based filtering technique is transformed into a predicate-based filtering technique, where we use DPR~\cite{karpukhin2020dense} to compute and rank the similarity between the candidate predicates and the query with the main entity removed.
%!TEX root = ../acl_latex.tex
\section{Experiments}

\subsection{Datasets}

To evaluate the performance of our method, we consider the widely used CRAG~\cite{yang2024crag} benchmark, where the KGs are accessed through API functions. We use the KG questions in CRAG for conducting experiments.

We further adapted our method to the Open domain question set of the Head-to-tail benchmark (Head2Tail)~\cite{sun2024head}, where the DBPedia KG is utilized, and the content is accessed through SPARQL queries to showcase the generalizability of our approach under different KGQA settings.

Both the CRAG and Head2Tail categorize the questions into head, torso, and tail based on the popularity of the \rev{topic entity}. We adopt the public validation set of CRAG as the training set and use the private testing set as the testing set. The statistics of the training and testing sets in the CRAG KG subset are presented in Table~\ref{tab:crag-stats} of Appendix~\ref{appendix:dataset}. As for the Head2Tail dataset, we randomly sampled 375 questions from each of the head, torso, and tail categories to formulate the testing set, while the rest can be treated as the training set. We present the detailed statistics of the datasets used in Appendix~\ref{appendix:dataset}.

\rev{We also included other SPARQL-based datasets used by the state-of-the-art approaches: QALD-10-en~\cite{Usbeck2023QALD10T}, WebQSP~\cite{yih2016value}, AdvHotpotQA~\cite{ye2022unreliability}, and CWQ~\cite{talmor-berant-2018-web}. }

\subsection{Baselines}
We compare our approach against the following baselines on both the CRAG and Head2Tail datasets. Details of baselines are in Appendix~\ref{appendix:baseline}.
\begin{itemize}[itemsep=0pt, topsep=0pt]
    \item {\em Base models:} GPT-4o~\cite{hurst2024gpt}, Llama-3.1-70B-Instruct~\cite{dubey2024llama}, GPT-4o (tool), and Llama-3.1-70B-Instruct (tool). The GPT-4o and Llama-3.1-70B-Instruct models, along with their tool-use variants (prompts are in Appendix~\ref{appendix:prompt-tool}), serve as representatives for closed-source and open-source LLM solutions. 
    \item {\em KDD Cup winners:} For CRAG, we include the top-2 KDD Cup CRAG winning solutions, db3~\cite{xia2024winning} and apex~\cite{ouyang2024revisiting}, which are state-of-the-art solutions on the CRAG benchmark. 
    \item {\em Other state-of-the-art models:} We consider the following state-of-the-art methods on the SPARQL-based Head2Tail dataset: WikiSP~\cite{xu2023fine}, StructGPT~\cite{jiang2023structgpt}, and ToG~\cite{sunthink}. Note that since CRAG does not support SPARQL querying, so we ran them only on Head2Tail. \rev{We further compared with the state-of-the-art results mentioned in ToG-2~\cite{ma2025thinkongraph} on QALD-10-en, WebQSP, AdvHotpotQA, and CWQ datasets. We did not evaluate ToG-2 on Head2Tail because ToG-2 is designed based on Wikidata and requires external pages to provide context, which is not compatible with the DBPeida KG used by Head2Tail. Besides, the extra information introduced by external pages would influence fairness of evaluation.}
\end{itemize}

Finally, for fair comparison with KDD Cup winning solutions, which may use 8B models, we also included KERAG (8B), where all modules are developed based on Llama-3.1-8B-Instruct.

\subsection{Evaluation}
To evaluate the performance of the approaches \rev{on CRAG and Head2Tail}, we adopted a similar auto-evaluation template used by CRAG (detailed in Appendix~\ref{appendix:prompt-auto}). The Llama-3.1-70B-Instruct model is employed as the critic. Following the settings in CRAG, we classify the answers into accurate, missing, and hallucination, corresponding to correct, missing, and incorrect answers. We report the accuracy ($A$), miss rate ($M$), hallucination rate ($H$), and additionally the {\em truthfulness} score ($T$), which is defined as $T=1*A+0*M-1*H$, penalizing the cases where the methods hallucinate.

\rev{For QALD-10-en, WebQSP, AdvHotpotQA, and CWQ datasets, we align with their respective evaluation methods and metrics. }

%\subsection{Configurations}
The configurations are presented in Appendix~\ref{appendix:config}.

\subsection{Experimental Results}

\label{sec:exp-results}

\begin{table}[t!]

  \caption{KERAG improves over state-of-the-art by 7.1\% on CRAG benchmark.}
  \vspace{-1em}
  \label{tab:baseline_crag}
\centering
\scriptsize
\begin{tabular}{lrrrrr}
\toprule 
 \textbf{Model} & \textbf{Accu.} & \textbf{Hall.} & \textbf{Miss.} & \textbf{Truth.} \\
\midrule
 GPT-4o   &     0.341               &                 0.090       &   0.569                &   0.251             \\
 Llama   &      0.306              &   0.080                     & 0.614                  & 0.227               \\
 GPT-4o (tool)   &    0.362                &      \textbf{0.047}                  &    0.592               &     0.315           \\
 Llama (tool)   &      0.220              &           0.057             &     0.723              &       0.163         \\
 apex~\cite{ouyang2024revisiting}   &       0.652             &    0.194                    &    0.154               &       0.458         \\
 db3~\cite{xia2024winning}   &         0.510           &  0.173                      &             0.317      &         0.337       \\
\cmidrule{1-5}
 %KERAG (8B)   &     0.713          &    0.208             &     0.080    &   0.505         \\
 KERAG   &     \textbf{0.732}               &                0.202        & \textbf{0.066}                  & \textbf{0.529}                            \\
\bottomrule
\end{tabular}
\vspace{-1em}
\end{table}

We present the experimental results for the CRAG dataset in Table~\ref{tab:baseline_crag}. We first notice that KERAG outperforms all baselines in CRAG, with an overall improvement of 7.1\% in terms of the truthfulness score. Specifically, we observe that the LLM baselines (GPT-4o and Llama) perform poorly (accuracy < 40\%) on the CRAG benchmark, which highlights the difficulty of the questions. In addition, although the tool-use variants of LLM have slightly lower hallucination rates, they fail to achieve high accuracy and have >55\% missing rates. We attribute the high missing rate of tool-use variants to the fact that direct function chaining is likely to propagate planning errors to the retrieval stage, leading to low recall of relevant KG content. 
%As a result, the LLM (tool) baselines exhibit a high miss rate along with low accuracy. 
The KDD Cup winning solutions significantly outperformed baseline LLM solutions, obtaining higher accuracy (40-65\%), at the price of much higher hallucination rate (\textasciitilde20\%). Our KERAG solution achieves much higher accuracy (>70\%), with very minor increase of hallucination rate; showcasing the superiority of multi-hop retrieval and filtering design by providing high recall of relevant content, thereby reducing the miss rate. %Moreover, the overall performance difference between KERAG (8B) and KERAG is below 3\%, demonstrating the robustness of our pipeline design. 
%(Compare the performance of KERAG against the above-listed baselines. )

%To validate the generalizability of KERAG on general SPARQL-based knowledge graph use cases, we adapt our KERAG pipeline to the open domain question set of the Head2Tail benchmark. 
We observed similar trends on the Head2Tail benchmark, shown in Table~\ref{tab:baseline_htt}. First, KERAG outperforms all the baselines in terms of overall accuracy and truthfulness scores, in particular significantly outperforming GPT-4o and Llama-3.1-70B-Instruct across all data splits. Furthermore, although we allow LLMs to write SPARQL queries to access the knowledge graph (tool variants), KERAG still shows overall truthfulness improvements of 10-15\% compared to GPT-4o (tool) and Llama-3.1-70B-Instruct (tool). Additionally, KERAG outperforms WikiSP, StructGPT, and ToG by \textasciitilde8\% on truthfulness scores, with lower hallucinations. \rev{We report that the retrieval recall of our approach is 0.952, while the recall of ToG is 0.844, validating the advantages of our neighborhood-based exploration against the path-based exploration of ToG.} Finally, we observed 91\% accuracy on this dataset, which mainly contains simple questions, showing that KERAG has mostly conquered simple-question answering.
%In summary, our experiments on Head2Tail showcase the generalizability of our approach to SPARQL-based knowledge graphs: KERAG achieves state-of-the-art performance on the Head2Tail benchmark.

\rev{We further present the results on QALD-10-en, WebQSP, AdvHotpotQA, and CWQ datasets in Table~\ref{tab:qald-results}: our approach outperforms SOTA methods on QALD-10-en, WebQSP, and AdvHotpotQA datasets and achieves comparable performance against SOTAs on CWQ dataset. }

\begin{table}[t]

  \caption{KERAG improves over state-of-the-art by 7\% on Head2Tail benchmark.}
  \vspace{-1em}
  \label{tab:baseline_htt}
\centering
\scriptsize
\begin{tabular}{lrrrrr}
\toprule 
 \textbf{Model} & \textbf{Accu.} & \textbf{Hall.} & \textbf{Miss.} & \textbf{Truth.} \\
 \midrule
	 GPT-4o   &     0.502               &       0.160                 &    0.338               &   0.342             \\
		 Llama   &    0.452                &  0.163                      &   0.385                &     0.290           \\
         GPT-4o (tool)   &       0.799             &    0.035                    &     0.166              &     0.764           \\
		 Llama (tool)   &            0.752        &              \textbf{0.031}          &         0.217          &         0.721       \\
		 WikiSP~\cite{xu2023fine}   &   0.858                 &  0.066                      &    0.076               &    0.782            \\
		  StructGPT~\cite{jiang2023structgpt}  &            0.895        &    0.105                    &       \textbf{0.000}          &      0.790          \\
          ToG~\cite{sunthink} & 0.833 & 0.063 & 0.104 & 0.770\\
          %ToG-2~\cite{ma2025thinkongraph} & \\
        \cmidrule{1-5}
		 KERAG   &  \textbf{0.908}                  & 0.049                       &   0.043                &     \textbf{0.860}           \\
\bottomrule
\end{tabular}
\vspace{-1em}
\end{table}

\begin{table}[t]

\centering
\scriptsize
\caption{\rev{Experimental results on QALD-10-en, WebQSP, AdvHotpotQA, and CWQ datasets. The prior SOTA include the best known methods on each dataset: $\alpha$:~\cite{borroto2022sparql}; $\beta$:~\cite{yu2022decaf}; $\gamma$:~\cite{lichain}; $\delta$:~\cite{das2021case}. The baseline performance is directly taken from ToG and ToG-2. ToG-2 additionally links to external Wikipedia pages to get context information.}}
\vspace{-1em}
\label{tab:qald-results}
\begin{tabular}{lllll}
\toprule
        \textbf{Model}             & \textbf{QALD-10-en} & \textbf{WebQSP} & \textbf{AdvHotpotQA} & \textbf{CWQ} \\    
\midrule
Prior SOTA & 0.454$^{\alpha}$ & 0.821$^{\beta}$ & 0.354$^{\gamma}$ & \textbf{0.704$^{\delta}$}\\
ToG & 0.502 & 0.762 & 0.263 & 0.695\\
ToG-2 & 0.541 & 0.811 & 0.429 & - \\
\midrule
KERAG & \textbf{0.558} & \textbf{0.843} & \textbf{0.461} & 0.702\\
\bottomrule
\end{tabular}

\end{table}

\begin{table}[t]

  \caption{Ablation Study on CRAG.}
  \vspace{-1em}
  \label{tab:ablation_crag}
\small
\centering
\begin{tabular}{llrrrrr}
\toprule 
 \textbf{Model} & \textbf{Accu.} & \textbf{Hall.} & \textbf{Miss.} & \textbf{Truth.} \\
\midrule
 KERAG   &     0.732               &                0.202        & 0.066                  & 0.529               \\
  - multihop  &  0.587                  &            0.112            &       0.301            &       0.474         \\
  - filter  &    0.721                &                    0.232    &  0.047                 &         0.490       \\
  - finetuning  &       0.626             &    0.173                    &                 0.201  &       0.453         \\
  - CoT  &    0.649                &                  0.348      &  0.003                 &         0.301       \\
\bottomrule
\end{tabular}

\end{table}

\subsection{Ablation Study}

We conducted an ablation study to explore the effectiveness of each component in KERAG: 1) -multihop: considers only one-hop neighbors in the retriever; 2) -filter: without performing the filtering step; 3) -finetuning: without fine-tuning the CoT-based summarizer; 4) -CoT: replaces the CoT-based summarizer with a plain tuned summarizer.

We observed that the overall truthfulness of these four variants decreases relative to KERAG by 7.4\%, 10.4\%, 14.4\%, 43.1\%. As expected, multi-hop retrieval increased missing rate by 25\%, with the price of increased hallucination rate (by 9\%). Filtering reduced hallucination rate by removing noises in the retrieval results. The Chain-of-Thought prompting of the summarizer and the fine-tuning process improved the quality most, increasing accuracy by 9\% and reducing hallucinations by 15\%; fine-tuning played an important role in reducing missing answers, especially for complex questions. %are the primary factors influencing the performance of KERAG. While the influence of the exhaustive retrieval and tool-based filtering may not be as significant as that of the fine-tuning and CoT of the summarizer, 
Combining the four designs maximizes the performance of KERAG.

We further explored the effect of multi-hop retrieval on KGs %with hard constraints 
by comparing the performance difference of our multi-hop retrieval strategy for one-hop, two-hop, and even n-hop on CRAG\footnote{In our case, the API schema of CRAG limits $n \le 3$.}. 
%In our case, the API schema of CRAG limits the maximum number of hops for API chaining to three. Therefore, we compare our exhaustive multihop retrieval against one-hop and two-hop strategies in CRAG and 
We present the results in Figure~\ref{fig:discussion} of the Appendix. We observe that the accuracy increased significantly (15\%) from one-hop to exhaustive retrieval; correspondingly, missing rate reduced significantly from 30\% for one-hop to 7\% for exhaustive. On the other hand, we also observe an increased hallucination rate from one-hop to two-hop (9\%), brought by retrieval noises, thereby validating the necessity of a proper content filter and a summarizer with high reasoning ability in the pipeline. %If the API schema did not limit the maximum number of hops, we could adopt the LLM-based soft constraint to determine the optimal number of hops of each query when using the KERAG pipeline.

%\subsection{Case Study}
%\label{sec:case}

%(Adapt the KERAG method to Head2Tail dataset and compare with the baselines, showcase the generalizability, suggested by Kai.)

\subsection{Robustness}

%Discuss head/torso/tail
We explore the robustness of KERAG on head/torso/tail categorization. As shown in Tables~\ref{tab:baseline_benchmarking_crag_full} and~\ref{tab:baseline_benchmarking_htt_full} in Appendix~\ref{appendix:exp-htt}, GPT-4o and Llama-3.1-70B-Instruct models present a significant decreasing trend in terms of accuracy and truthfulness going from head to torso to tail categories, even with KG-based RAG. In contrast, as shown in Table~\ref{tab:kerag-htt-results}, KERAG presents a relatively stable performance, narrowing down the gaps and demonstrating robustness. Interestingly, on Head2Tail KERAG has lower performance on head entities, because of challenges like knowledge overloading on head entities and ambiguous entity names.

\rev{To further analyze the generalizability of KERAG, we adapt our KERAG approach to a multi-entity settings, where the planner is asked to identify a list of entities instead of a single entity and hops from multiple topic entities to retrieve relevant information. In this way, our solution is generalized to the cases where retrieving information of multiple entities is necessary. In order to evaluate the performance of the generalized solution on multi-entity cases, we experiment the adapted KERAG model on comparison questions, which is the only question type that involves multiple entities in CRAG. As shown in Table~\ref{tab:multi-entity}, before applying the adaptation the KERAG model achieve truthfulness score (accuracy-hallucination rate) and accuracy of 0.7 and 0.817 respectively. We achieve truthfulness score and accuracy of 0.75 and 0.85 respectively with multi-entity adaptation, validating the feasibility and effectiveness of the adaptation.}

\begin{table}[hbtp]

  \caption{\rev{Multi-entity Generalizability.}}
  \label{tab:multi-entity}
  \vspace{-1em}
\small
\centering
\begin{tabular}{llrrrrr}
\toprule 
 \textbf{Model} & \textbf{A} & \textbf{H} & \textbf{M} & \textbf{T} \\
\midrule
 KERAG (single)   &     0.817               &                0.117        & 0.066                  & 0.700               \\
  KERAG (multiple)  &  0.850                  &            0.100            &       0.050            &       0.750         \\
\bottomrule
\end{tabular}

\end{table}

%Different LLM backbones are evaluated to show the robustness of KERAG in Appendix~\ref{sec:exp-backbone}.}
\rev{We present the experimental results by replacing the LLM backbones of KERAG in Table~\ref{tab:backbone}. We observe that the performance drops slightly when we replace the Llama-3.1 70B with Llama-3.1 8B. We additionally replace the backbone with DeepSeek-chat model (DeepSeek-V3) to see if a more advanced model brings performance uplift on CRAG dataset. The performance of KERAG (DeepSeek) further boosts the performance of KERAG (70B) by 2.4\% and 11.8\% in terms of overall accuracy and truthfulness scores (accuracy-hallucination rate). We believe that an insight is that advanced LLMs would reduce the hallucination risk in the planning step and thus bring performance uplift.}

\begin{table}[hbtp]

  \caption{\rev{Robustness of LLM Backbones.}}
  \vspace{-1em}
  \label{tab:backbone}
\small
\centering
\begin{tabular}{llrrrrr}
\toprule 
 \textbf{Model} & \textbf{A} & \textbf{H} & \textbf{M} & \textbf{T} \\
\midrule
 KERAG (8B)   &     0.713               &     0.208                   &    0.080            &     0.505         \\
  KERAG (70B)  &    0.732                &        0.202              &    0.066            &  0.529            \\
  KERAG (DeepSeek)  &     0.756          &       0.109             &  0.135                &     0.647         \\
\bottomrule
\end{tabular}

\end{table}
% reliability of each stage
\rev{We perform the error analysis on each component of KERAG as follows: For CRAG, The overall recall of content retrieval is 0.967 and the retrieval miss rate is 0.033. The summarization accuracy is 0.757, indicating that the main error lies in the summarization step. For Head2Tail, the entity linking accuracy is 0.957, the retrieval recall of the answer entity is 0.952, and the overall accuracy is 0.908, which indicates that the main errors occur in the entity linking and summarization.}

\begin{table}[hbtp]

  \caption{Head/torso/tail results of KERAG.}  
  \vspace{-1em}
  \label{tab:kerag-htt-results}
\centering
\small
\begin{tabular}{lrrrrr}
\toprule 
 \textbf{Data} & \textbf{A} & \textbf{H} & \textbf{M} & \textbf{T} \\
\midrule
CRAG-head    &     0.754             &  0.219                      &   0.027                &     0.535          \\
%\midrule
CRAG-torso    &     0.734             &     0.207                   &   0.059      &     0.527        \\
%\midrule
CRAG-tail    &     0.707           &    0.181                    &   0.112              &  0.527        \\
%\midrule
CRAG-overall    &     0.732             &                0.202        & 0.066                 & 0.529                            \\
\midrule
Head2Tail-head  &   0.883            &       0.067               &     0.051          &    0.816            \\
%\midrule
Head2Tail-torso &    0.917          &        0.045           &        0.037         &   0.872          \\
%\midrule
Head2Tail-tail     & 0.925                & 0.035                  &     0.040           &   0.891          \\
%\midrule
Head2Tail-overall   &  0.908                  & 0.049                       &   0.043                &     0.860                                   \\
\bottomrule
\end{tabular}

\end{table}

\subsection{Latency}
\label{sec:exp-efficiency}

\rev{We report the mean inference time of KERAG against the baselines in Table~\ref{tab:efficiency}. Specifically, although our approach presents a slightly larger latency than apex on CRAG, it demonstrates a smaller latency than StructGPT and ToG on Head2Tail dataset. In summary, we believe the overall latency of KERAG is acceptable.}

\begin{table}[hbtp]
  \caption{\rev{Mean Inference Time of KERAG and baselines.}}
  \label{tab:efficiency}
  \vspace{-1em}
\small
\centering
\begin{tabular}{lrr}
\toprule 
 \textbf{Model} & \textbf{CRAG} & \textbf{Head2Tail} \\
\midrule
 apex~\cite{ouyang2024revisiting} & 6.96s & - \\
 StructGPT~\cite{jiang2023structgpt} & - & 4.46s \\
 ToG~\cite{sunthink} & - & 4.96s \\
 KERAG & 8.55s & 3.18s \\
\bottomrule
\end{tabular}

\end{table}

%!TEX root = ../acl_latex.tex
\section{Conclusion}
\vspace{-0.5em}
In conclusion, our proposed KERAG pipeline addresses key challenges in KG-based RAG by leveraging entity-level information, significantly reducing reliance on triple-level semantic parsing. By integrating a novel data generation scheme for fine-tuning chain-of-thought summarizers, KERAG enhances the ability to handle complex queries effectively. Its flexibility across various knowledge graphs, including those accessed via SPARQL or APIs, demonstrates broad applicability and generalizability. Experimental results confirm that KERAG outperforms existing state-of-the-art methods, making it a promising advancement in KGQA.

%\newpage
\section*{Limitations}
Although our proposed KERAG approach demonstrates remarkable performance on both API-based and SPARQL-based knowledge graphs, we must acknowledge that the evaluation only covered CRAG, Head2Tail, QALD-10-en, WebQSP, AdvHotpotQA, and CWQ datasets. It is unknown how our proposed approach would perform in other KGs with different characteristics or domains. Furthermore, due to the multi-stage design of our approach, we have to admit that there may exist potential risks regarding error propagation in the pipeline. Our error analysis indicates that the main error might occur in entity linking and our current method does not focus on deploying a specific entity-linking mechanism. We believe that existing state-of-the-art entity linking approaches can be used as a pre-step to effectively integrate with our approach to improve the overall performance. We would like to point out that these limitations do not diminish the importance of our work; instead, they highlight potential areas for future research and enhancement.

%Although our proposed KERAG approach demonstrates remarkable performance on both API-based and SPARQL-based knowledge graphs, we must acknowledge that there exist limitations in the current method: Most of the base models we used belong to the Llama-3.1 family. Undoubtedly, more advanced large language models (LLMs) should achieve better performance. However, we believe that our framework is compatible with advanced LLMs. Interested researchers can replace the base model to verify the performance of other LLMs within the KERAG framework.

\section*{Ethics Statement}
In this section, we discuss the ethical considerations of our work from two aspects: 1) The knowledge graphs (KGs) utilized in our experiments include real-world named entities, such as individuals and organizations, along with their publicly accessible information, which cannot be anonymized when evaluating QA performance. The datasets referenced in this work~\cite{yang2024crag, sun2024head, Usbeck2023QALD10T, yih2016value, ye2022unreliability, talmor-berant-2018-web} were collected from publicly available online sources and have been extensively used in prior research. However, it is crucial to note that the use of these public datasets does not guarantee the absence of privacy violations. 2) Environmental Impact: The process of training and inferring models leads to energy consumption and carbon emissions. 

\section*{Acknowledgements}

Lei Chen's work is partially supported by National Key Research and Development Program of China Grant No. 2023YFF0725100, National Science Foundation of China (NSFC) under Grant No. U22B2060, Guangdong-Hong Kong Technology Innovation Joint Funding Scheme Project No. 2024A0505040012, the Hong Kong RGC GRF Project 16213620, RIF Project R6020-19, AOE Project AoE/E-603/18, Theme-based project TRS T41-603/20R, CRF Project C2004-21G, Key Areas Special Project of Guangdong Provincial  Universities 2024ZDZX1006, Guangdong Province Science and Technology Plan Project 2023A0505030011, Guangzhou municipality big data intelligence key lab, 2023A03J0012, Hong Kong ITC ITF grants MHX/078/21 and PRP/004/22FX, Zhujiang scholar program 2021JC02X170, Microsoft Research Asia Collaborative Research Grant, HKUST-Webank joint research lab and 2023 HKUST Shenzhen-Hong Kong Collaborative Innovation Institute Green Sustainability Special Fund, from Shui On Xintiandi and the InnoSpace GBA.

Nan Tang's work is partially supported by Guangdong provincial project 2023CX10X008.

\bibliography{custom}

\begin{thebibliography}{52}
\providecommand{\natexlab}[1]{#1}

\bibitem[{Borroto et~al.(2022)Borroto, Ricca, Cuteri, and Barbara}]{borroto2022sparql}
Manuel Borroto, Francesco Ricca, Bernardo Cuteri, and Vito Barbara. 2022.
\newblock \href {https://ceur-ws.org/Vol-3196/paper3.pdf} {Sparql-qa enters the qald challenge}.
\newblock In \emph{Proceedings of the 7th Natural Language Interfaces for the Web of Data (NLIWoD) co-located with the 19th European Semantic Web Conference, Hersonissos, Greece}, volume 3196, pages 25--31.

\bibitem[{Chen et~al.(2025)Chen, Song, Shum, Lin, Sheng, and Qu}]{chen2025unmasking}
Zixin Chen, Sicheng Song, Kashun Shum, Yanna Lin, Rui Sheng, and Huamin Qu. 2025.
\newblock Unmasking deceptive visuals: Benchmarking multimodal large language models on misleading chart question answering.
\newblock \emph{arXiv preprint arXiv:2503.18172}.

\bibitem[{Das et~al.(2021)Das, Zaheer, Thai, Godbole, Perez, Lee, Tan, Polymenakos, and Mccallum}]{das2021case}
Rajarshi Das, Manzil Zaheer, Dung Thai, Ameya Godbole, Ethan Perez, Jay-Yoon Lee, Lizhen Tan, Lazaros Polymenakos, and Andrew Mccallum. 2021.
\newblock \href {https://doi.org/10.18653/v1/2021.emnlp-main.755} {Case-based reasoning for natural language queries over knowledge bases}.
\newblock In \emph{Proceedings of the 2021 Conference on Empirical Methods in Natural Language Processing}, pages 9594--9611.

\bibitem[{Dong(2024)}]{Dong2024QA}
Xin~Luna Dong. 2024.
\newblock \href {https://doi.org/10.1145/3626246.3655999} {The journey to a knowledgeable assistant with retrieval-augmented generation (rag)}.
\newblock In \emph{Companion of the 2024 International Conference on Management of Data}, SIGMOD/PODS '24, page~3, New York, NY, USA. Association for Computing Machinery.

\bibitem[{Dong et~al.(2014)Dong, Gabrilovich, Heitz, Horn, Murphy, Sun, and Zhang}]{dong2014data}
Xin~Luna Dong, Evgeniy Gabrilovich, Geremy Heitz, Wilko Horn, Kevin Murphy, Shaohua Sun, and Wei Zhang. 2014.
\newblock \href {https://doi.org/10.14778/2732951.2732962} {From data fusion to knowledge fusion}.
\newblock \emph{Proceedings of the VLDB Endowment}, 7(10):881--892.

\bibitem[{Dubey et~al.(2024)Dubey, Jauhri, Pandey, Kadian, Al-Dahle, Letman, Mathur, Schelten, Yang, Fan et~al.}]{dubey2024llama}
Abhimanyu Dubey, Abhinav Jauhri, Abhinav Pandey, Abhishek Kadian, Ahmad Al-Dahle, Aiesha Letman, Akhil Mathur, Alan Schelten, Amy Yang, Angela Fan, and 1 others. 2024.
\newblock \href {https://doi.org/10.48550/arXiv.2407.21783} {The llama 3 herd of models}.
\newblock \emph{arXiv preprint arXiv:2407.21783}.

\bibitem[{Fan et~al.(2024)Fan, Ding, Ning, Wang, Li, Yin, Chua, and Li}]{fan2024survey}
Wenqi Fan, Yujuan Ding, Liangbo Ning, Shijie Wang, Hengyun Li, Dawei Yin, Tat-Seng Chua, and Qing Li. 2024.
\newblock \href {https://doi.org/10.1145/3637528.3671470} {A survey on rag meeting llms: Towards retrieval-augmented large language models}.
\newblock In \emph{Proceedings of the 30th ACM SIGKDD Conference on Knowledge Discovery and Data Mining}, pages 6491--6501.

\bibitem[{Gu et~al.(2023)Gu, Deng, and Su}]{gu2023don}
Yu~Gu, Xiang Deng, and Yu~Su. 2023.
\newblock \href {https://doi.org/10.18653/v1/2023.acl-long.270} {Don’t generate, discriminate: A proposal for grounding language models to real-world environments}.
\newblock In \emph{Proceedings of the 61st Annual Meeting of the Association for Computational Linguistics (Volume 1: Long Papers)}, pages 4928--4949.

\bibitem[{Gu and Su(2022)}]{gu2022arcaneqa}
Yu~Gu and Yu~Su. 2022.
\newblock \href {https://aclanthology.org/2022.coling-1.148/} {Arcaneqa: Dynamic program induction and contextualized encoding for knowledge base question answering}.
\newblock In \emph{Proceedings of the 29th International Conference on Computational Linguistics}, pages 1718--1731.

\bibitem[{He et~al.(2021)He, Lan, Jiang, Zhao, and Wen}]{he2021improving}
Gaole He, Yunshi Lan, Jing Jiang, Wayne~Xin Zhao, and Ji-Rong Wen. 2021.
\newblock \href {https://doi.org/10.1145/3437963.3441753} {Improving multi-hop knowledge base question answering by learning intermediate supervision signals}.
\newblock In \emph{Proceedings of the 14th ACM international conference on web search and data mining}, pages 553--561.

\bibitem[{Hogan et~al.(2025)Hogan, Dong, Vrande{\v{c}}i{\'c}, and Weikum}]{hogan2025large}
Aidan Hogan, Xin~Luna Dong, Denny Vrande{\v{c}}i{\'c}, and Gerhard Weikum. 2025.
\newblock \href {https://doi.org/10.48550/arXiv.2501.06699} {Large language models, knowledge graphs and search engines: A crossroads for answering users' questions}.
\newblock \emph{arXiv preprint arXiv:2501.06699}.

\bibitem[{Hu et~al.(2021)Hu, Shen, Wallis, Allen-Zhu, Li, Wang, Wang, and Chen}]{hu2021lora}
Edward~J Hu, Yelong Shen, Phillip Wallis, Zeyuan Allen-Zhu, Yuanzhi Li, Shean Wang, Lu~Wang, and Weizhu Chen. 2021.
\newblock \href {https://doi.org/10.48550/arXiv.2106.09685} {Lora: Low-rank adaptation of large language models}.
\newblock \emph{arXiv preprint arXiv:2106.09685}.

\bibitem[{Hu et~al.(2022)Hu, Wu, Shu, and Qu}]{hu2022logical}
Xixin Hu, Xuan Wu, Yiheng Shu, and Yuzhong Qu. 2022.
\newblock \href {https://aclanthology.org/2022.coling-1.145/} {Logical form generation via multi-task learning for complex question answering over knowledge bases}.
\newblock In \emph{Proceedings of the 29th International Conference on Computational Linguistics}, pages 1687--1696.

\bibitem[{Hurst et~al.(2024)Hurst, Lerer, Goucher, Perelman, Ramesh, Clark, Ostrow, Welihinda, Hayes, Radford et~al.}]{hurst2024gpt}
Aaron Hurst, Adam Lerer, Adam~P Goucher, Adam Perelman, Aditya Ramesh, Aidan Clark, AJ~Ostrow, Akila Welihinda, Alan Hayes, Alec Radford, and 1 others. 2024.
\newblock \href {https://doi.org/10.48550/arXiv.2410.21276} {Gpt-4o system card}.
\newblock \emph{arXiv preprint arXiv:2410.21276}.

\bibitem[{Jiang et~al.(2023)Jiang, Zhou, Dong, Ye, Zhao, and Wen}]{jiang2023structgpt}
Jinhao Jiang, Kun Zhou, Zican Dong, Keming Ye, Wayne~Xin Zhao, and Ji-Rong Wen. 2023.
\newblock \href {https://doi.org/10.18653/v1/2023.emnlp-main.574} {Structgpt: A general framework for large language model to reason over structured data}.
\newblock In \emph{Proceedings of the 2023 Conference on Empirical Methods in Natural Language Processing}, pages 9237--9251.

\bibitem[{Karpukhin et~al.(2020)Karpukhin, Oguz, Min, Lewis, Wu, Edunov, Chen, and Yih}]{karpukhin2020dense}
Vladimir Karpukhin, Barlas Oguz, Sewon Min, Patrick Lewis, Ledell Wu, Sergey Edunov, Danqi Chen, and Wen-tau Yih. 2020.
\newblock \href {https://doi.org/10.18653/v1/2020.emnlp-main.550} {Dense passage retrieval for open-domain question answering}.
\newblock In \emph{Proceedings of the 2020 Conference on Empirical Methods in Natural Language Processing (EMNLP)}, pages 6769--6781.

\bibitem[{Lan and Jiang(2020)}]{lan2020query}
Yunshi Lan and Jing Jiang. 2020.
\newblock \href {https://doi.org/10.18653/v1/2020.acl-main.91} {Query graph generation for answering multi-hop complex questions from knowledge bases}.
\newblock In \emph{Proceedings of the 58th Annual Meeting of the Association for Computational Linguistics}, pages 969--974.

\bibitem[{Lan et~al.(2019)Lan, Wang, and Jiang}]{lanknowledge}
Yunshi Lan, Shuohang Wang, and Jing Jiang. 2019.
\newblock \href {https://www.ijcai.org/proceedings/2019/0701.pdf} {Knowledge base question answering with topic units.(2019)}.
\newblock In \emph{Proceedings of the Twenty-Eighth International Joint Conference on Artificial Intelligence}, pages 5046--5052.

\bibitem[{Li et~al.(2025)Li, Jung, Chen, Wang, Wang, Qu, and Lau}]{li2025pipe}
Haobo Li, Eunseo Jung, Zixin Chen, Zhaowei Wang, Yueya Wang, Huamin Qu, and Alexis Kai~Hon Lau. 2025.
\newblock Pipe: Physics-informed position encoding for alignment of satellite images and time series.
\newblock \emph{arXiv preprint arXiv:2506.14786}.

\bibitem[{Li et~al.(2024{\natexlab{a}})Li, Wang, Wang, Wang, Lau, and Qu}]{li2024cllmate}
Haobo Li, Zhaowei Wang, Jiachen Wang, YueYa Wang, Alexis Kai~Hon Lau, and Huamin Qu. 2024{\natexlab{a}}.
\newblock Cllmate: A multimodal benchmark for weather and climate events forecasting.
\newblock \emph{arXiv preprint arXiv:2409.19058}.

\bibitem[{Li et~al.(2022)Li, Sun, Hsu, Li, and Wong}]{li2022fake}
Tianle Li, Yushi Sun, Shang-ling Hsu, Yanjia Li, and Raymond Chi-Wing Wong. 2022.
\newblock \href {https://arxiv.org/pdf/2205.03100} {Fake news detection with heterogeneous transformer}.
\newblock \emph{arXiv preprint arXiv:2205.03100}.

\bibitem[{Li et~al.(2024{\natexlab{b}})Li, Zhao, Chia, Ding, Joty, Poria, and Bing}]{lichain}
Xingxuan Li, Ruochen Zhao, Yew~Ken Chia, Bosheng Ding, Shafiq Joty, Soujanya Poria, and Lidong Bing. 2024{\natexlab{b}}.
\newblock \href {https://openreview.net/pdf?id=cPgh4gWZlz} {Chain-of-knowledge: Grounding large language models via dynamic knowledge adapting over heterogeneous sources}.
\newblock In \emph{The Twelfth International Conference on Learning Representations}.

\bibitem[{Luo et~al.(2018)Luo, Lin, Luo, and Zhu}]{luo2018knowledge}
Kangqi Luo, Fengli Lin, Xusheng Luo, and Kenny Zhu. 2018.
\newblock \href {https://doi.org/10.18653/v1/D18-1242} {Knowledge base question answering via encoding of complex query graphs}.
\newblock In \emph{Proceedings of the 2018 conference on empirical methods in natural language processing}, pages 2185--2194.

\bibitem[{Ma et~al.(2025)Ma, Xu, Jiang, Li, Qu, Yang, Mao, and Guo}]{ma2025thinkongraph}
Shengjie Ma, Chengjin Xu, Xuhui Jiang, Muzhi Li, Huaren Qu, Cehao Yang, Jiaxin Mao, and Jian Guo. 2025.
\newblock \href {https://openreview.net/forum?id=oFBu7qaZpS} {Think-on-graph 2.0: Deep and faithful large language model reasoning with knowledge-guided retrieval augmented generation}.
\newblock In \emph{The Thirteenth International Conference on Learning Representations}.

\bibitem[{Mavromatis and Karypis(2022)}]{mavromatis2022rearev}
Costas Mavromatis and George Karypis. 2022.
\newblock \href {https://doi.org/10.18653/v1/2022.findings-emnlp.181} {Rearev: Adaptive reasoning for question answering over knowledge graphs}.
\newblock In \emph{Findings of the Association for Computational Linguistics: EMNLP 2022}, pages 2447--2458.

\bibitem[{Ni et~al.(2025)Ni, Liu, Wang, Lei, Zhao, Cheng, Zeng, Dong, Xia, Kenthapadi et~al.}]{ni2025towards}
Bo~Ni, Zheyuan Liu, Leyao Wang, Yongjia Lei, Yuying Zhao, Xueqi Cheng, Qingkai Zeng, Luna Dong, Yinglong Xia, Krishnaram Kenthapadi, and 1 others. 2025.
\newblock \href {https://doi.org/10.48550/arXiv.2502.06872} {Towards trustworthy retrieval augmented generation for large language models: A survey}.
\newblock \emph{arXiv preprint arXiv:2502.06872}.

\bibitem[{Oguz et~al.(2022)Oguz, Chen, Karpukhin, Peshterliev, Okhonko, Schlichtkrull, Gupta, Mehdad, and Yih}]{oguz2022unik}
Barlas Oguz, Xilun Chen, Vladimir Karpukhin, Stan Peshterliev, Dmytro Okhonko, Michael Schlichtkrull, Sonal Gupta, Yashar Mehdad, and Scott Yih. 2022.
\newblock \href {https://doi.org/10.18653/v1/2022.findings-naacl.115} {Unik-qa: Unified representations of structured and unstructured knowledge for open-domain question answering}.
\newblock In \emph{Findings of the Association for Computational Linguistics: NAACL 2022}, pages 1535--1546.

\bibitem[{Ouyang et~al.(2024)Ouyang, Luo, Cheng, Wang, Yu, Liu, and Chen}]{ouyang2024revisiting}
Jie Ouyang, Yucong Luo, Mingyue Cheng, Daoyu Wang, Shuo Yu, Qi~Liu, and Enhong Chen. 2024.
\newblock \href {https://openreview.net/forum?id=PUzLjWIgqC} {Revisiting the solution of meta kdd cup 2024: Crag}.
\newblock \emph{arXiv preprint arXiv:2409.15337}.

\bibitem[{Saxena et~al.(2020)Saxena, Tripathi, and Talukdar}]{saxena2020improving}
Apoorv Saxena, Aditay Tripathi, and Partha Talukdar. 2020.
\newblock \href {https://doi.org/10.18653/v1/2020.acl-main.412} {Improving multi-hop question answering over knowledge graphs using knowledge base embeddings}.
\newblock In \emph{Proceedings of the 58th annual meeting of the association for computational linguistics}, pages 4498--4507.

\bibitem[{Sen et~al.(2021)Sen, Oliya, and Saffari}]{sen2021expanding}
Priyanka Sen, Armin Oliya, and Amir Saffari. 2021.
\newblock \href {https://doi.org/10.18653/v1/2021.emnlp-main.694} {Expanding end-to-end question answering on differentiable knowledge graphs with intersection}.
\newblock In \emph{Proceedings of the 2021 Conference on Empirical Methods in Natural Language Processing}, pages 8805--8812.

\bibitem[{Shi et~al.(2021)Shi, Cao, Hou, Li, and Zhang}]{shi2021transfernet}
Jiaxin Shi, Shulin Cao, Lei Hou, Juanzi Li, and Hanwang Zhang. 2021.
\newblock \href {https://doi.org/10.18653/v1/2021.emnlp-main.341} {Transfernet: An effective and transparent framework for multi-hop question answering over relation graph}.
\newblock In \emph{Proceedings of the 2021 Conference on Empirical Methods in Natural Language Processing}, pages 4149--4158.

\bibitem[{Shu et~al.(2022)Shu, Yu, Li, Karlsson, Ma, Qu, and Lin}]{shu2022tiara}
Yiheng Shu, Zhiwei Yu, Yuhan Li, B{\"o}rje Karlsson, Tingting Ma, Yuzhong Qu, and Chin-Yew Lin. 2022.
\newblock \href {https://doi.org/10.18653/v1/2022.emnlp-main.555} {Tiara: Multi-grained retrieval for robust question answering over large knowledge base}.
\newblock In \emph{Proceedings of the 2022 Conference on Empirical Methods in Natural Language Processing}, pages 8108--8121.

\bibitem[{Sun et~al.(2024{\natexlab{a}})Sun, Xu, Tang, Wang, Lin, Gong, Ni, Shum, and Guo}]{sunthink}
Jiashuo Sun, Chengjin Xu, Lumingyuan Tang, Saizhuo Wang, Chen Lin, Yeyun Gong, Lionel Ni, Heung-Yeung Shum, and Jian Guo. 2024{\natexlab{a}}.
\newblock \href {https://openreview.net/forum?id=nnVO1PvbTv} {Think-on-graph: Deep and responsible reasoning of large language model on knowledge graph}.
\newblock In \emph{The Twelfth International Conference on Learning Representations}.

\bibitem[{Sun et~al.(2024{\natexlab{b}})Sun, Xu, Zha, Liu, and Dong}]{sun2024head}
Kai Sun, Yifan Xu, Hanwen Zha, Yue Liu, and Xin~Luna Dong. 2024{\natexlab{b}}.
\newblock \href {https://doi.org/10.18653/v1/2024.naacl-long.18} {Head-to-tail: How knowledgeable are large language models (llms)? aka will llms replace knowledge graphs?}
\newblock In \emph{Proceedings of the 2024 Conference of the North American Chapter of the Association for Computational Linguistics: Human Language Technologies (Volume 1: Long Papers)}, pages 311--325.

\bibitem[{Sun et~al.(2025)Sun, Sun, Yang, and Tang}]{sun2025knowledge}
Yushi Sun, Kai Sun, Xiao Yang, and Nan Tang. 2025.
\newblock \href {https://ebooks.iospress.nl/volumearticle/72296} {Knowledge internalized in llms}.
\newblock In \emph{Handbook on Neurosymbolic AI and Knowledge Graphs}, pages 230--255. IOS Press.

\bibitem[{Sun et~al.(2024{\natexlab{c}})Sun, Wang, Cheng, Zheng, Chen, and Yin}]{sun2024cross}
Yushi Sun, Jiachuan Wang, Peng Cheng, Libin Zheng, Lei Chen, and Jian Yin. 2024{\natexlab{c}}.
\newblock \href {https://doi.org/10.1109/ICDE60146.2024.00026} {Cross-domain-aware worker selection with training for crowdsourced annotation}.
\newblock In \emph{2024 IEEE 40th International Conference on Data Engineering (ICDE)}, pages 249--262. IEEE.

\bibitem[{Sun et~al.(2023)Sun, Xin, and Chen}]{sun2023reca}
Yushi Sun, Hao Xin, and Lei Chen. 2023.
\newblock \href {https://doi.org/10.14778/3583140.3583149} {Reca: Related tables enhanced column semantic type annotation framework}.
\newblock \emph{Proceedings of the VLDB Endowment}, 16(6):1319--1331.

\bibitem[{Sun et~al.(2024{\natexlab{d}})Sun, Xin, Sun, Xu, Yang, Dong, Tang, and Chen}]{sun2024large}
Yushi Sun, Hao Xin, Kai Sun, Yifan~Ethan Xu, Xiao Yang, Xin~Luna Dong, Nan Tang, and Lei Chen. 2024{\natexlab{d}}.
\newblock \href {https://doi.org/10.14778/3681954.3681973} {Are large language models a good replacement of taxonomies?}
\newblock \emph{Proceedings of the VLDB Endowment}, 17(11):2919--2932.

\bibitem[{Talmor and Berant(2018)}]{talmor-berant-2018-web}
Alon Talmor and Jonathan Berant. 2018.
\newblock \href {https://doi.org/10.18653/v1/N18-1059} {The web as a knowledge-base for answering complex questions}.
\newblock In \emph{Proceedings of the 2018 Conference of the North {A}merican Chapter of the Association for Computational Linguistics: Human Language Technologies, Volume 1 (Long Papers)}, pages 641--651, New Orleans, Louisiana. Association for Computational Linguistics.

\bibitem[{Usbeck et~al.(2023)Usbeck, Yan, Perevalov, Jiang, Schulz, Kraft, M{\"o}ller, Huang, Reineke, Ngomo, Saleem, and Both}]{Usbeck2023QALD10T}
Ricardo Usbeck, Xi~Yan, Aleksandr Perevalov, Longquan Jiang, Julius Schulz, Angelie Kraft, Cedric M{\"o}ller, Junbo Huang, Jan Reineke, Axel-Cyrille~Ngonga Ngomo, Muhammad Saleem, and Andreas Both. 2023.
\newblock \href {https://api.semanticscholar.org/CorpusID:265577096} {Qald-10 – the 10th challenge on question answering over linked data}.
\newblock \emph{Semantic Web}.

\bibitem[{Wang et~al.(2024)Wang, Huey, Sheng, Mehta, and Wang}]{wang2024scidasynth}
Xingbo Wang, Samantha~L Huey, Rui Sheng, Saurabh Mehta, and Fei Wang. 2024.
\newblock Scidasynth: Interactive structured knowledge extraction and synthesis from scientific literature with large language model.
\newblock \emph{arXiv preprint arXiv:2404.13765}.

\bibitem[{Xia et~al.(2024)Xia, Chen, and Gao}]{xia2024winning}
Yikuan Xia, Jiazun Chen, and Jun Gao. 2024.
\newblock \href {https://openreview.net/forum?id=oWNPeoP1uC} {Winning solution for meta kdd cup'24}.
\newblock \emph{arXiv preprint arXiv:2410.00005}.

\bibitem[{Xie et~al.(2022)Xie, Wu, Shi, Zhong, Scholak, Yasunaga, Wu, Zhong, Yin, Wang et~al.}]{xie2022unifiedskg}
Tianbao Xie, Chen~Henry Wu, Peng Shi, Ruiqi Zhong, Torsten Scholak, Michihiro Yasunaga, Chien-Sheng Wu, Ming Zhong, Pengcheng Yin, Sida~I Wang, and 1 others. 2022.
\newblock \href {https://doi.org/10.18653/v1/2022.emnlp-main.39} {Unifiedskg: Unifying and multi-tasking structured knowledge grounding with text-to-text language models}.
\newblock In \emph{Proceedings of the 2022 Conference on Empirical Methods in Natural Language Processing}, pages 602--631.

\bibitem[{Xu et~al.(2023)Xu, Liu, Culhane, Pertseva, Wu, Semnani, and Lam}]{xu2023fine}
Silei Xu, Shicheng Liu, Theo Culhane, Elizaveta Pertseva, Meng-Hsi Wu, Sina Semnani, and Monica Lam. 2023.
\newblock \href {https://doi.org/10.18653/v1/2023.emnlp-main.353} {Fine-tuned llms know more, hallucinate less with few-shot sequence-to-sequence semantic parsing over wikidata}.
\newblock In \emph{Proceedings of the 2023 Conference on Empirical Methods in Natural Language Processing}, pages 5778--5791.

\bibitem[{Yang et~al.(2024)Yang, Sun, Xin, Sun, Bhalla, Chen, Choudhary, Gui, Jiang, Jiang, Kong, Moran, Wang, Xu, Yan, Yang, Yuan, Zha, Tang, Chen, Scheffer, Liu, Shah, Wanga, Kumar, tau Yih, and Dong}]{yang2024crag}
Xiao Yang, Kai Sun, Hao Xin, Yushi Sun, Nikita Bhalla, Xiangsen Chen, Sajal Choudhary, Rongze~Daniel Gui, Ziran~Will Jiang, Ziyu Jiang, Lingkun Kong, Brian Moran, Jiaqi Wang, Yifan~Ethan Xu, An~Yan, Chenyu Yang, Eting Yuan, Hanwen Zha, Nan Tang, and 8 others. 2024.
\newblock \href {https://arxiv.org/abs/2406.04744} {Crag -- comprehensive rag benchmark}.
\newblock \emph{arXiv preprint arXiv:2406.04744}.

\bibitem[{Ye and Durrett(2022)}]{ye2022unreliability}
Xi~Ye and Greg Durrett. 2022.
\newblock \href {https://openreview.net/pdf?id=Bct2f8fRd8S} {The unreliability of explanations in few-shot prompting for textual reasoning}.
\newblock \emph{Advances in neural information processing systems}, 35:30378--30392.

\bibitem[{Yih et~al.(2015)Yih, Chang, He, and Gao}]{yih2015semantic}
Scott Wen-tau Yih, Ming-Wei Chang, Xiaodong He, and Jianfeng Gao. 2015.
\newblock \href {https://aclanthology.org/P15-1128.pdf} {Semantic parsing via staged query graph generation: Question answering with knowledge base}.
\newblock In \emph{Proceedings of the Joint Conference of the 53rd Annual Meeting of the ACL and the 7th International Joint Conference on Natural Language Processing of the AFNLP}.

\bibitem[{Yih et~al.(2016)Yih, Richardson, Meek, Chang, and Suh}]{yih2016value}
Wen-tau Yih, Matthew Richardson, Christopher Meek, Ming-Wei Chang, and Jina Suh. 2016.
\newblock \href {https://aclanthology.org/P16-2033.pdf} {The value of semantic parse labeling for knowledge base question answering}.
\newblock In \emph{Proceedings of the 54th Annual Meeting of the Association for Computational Linguistics (Volume 2: Short Papers)}, pages 201--206.

\bibitem[{Yu et~al.(2022)Yu, Zhang, Ng, Zhu, Li, Wang, Hu, Wang, Wang, and Xiang}]{yu2022decaf}
Donghan Yu, Sheng Zhang, Patrick Ng, Henghui Zhu, Alexander~Hanbo Li, Jun Wang, Yiqun Hu, William~Yang Wang, Zhiguo Wang, and Bing Xiang. 2022.
\newblock \href {https://openreview.net/pdf?id=XHc5zRPxqV9} {Decaf: Joint decoding of answers and logical forms for question answering over knowledge bases}.
\newblock In \emph{The Eleventh International Conference on Learning Representations}.

\bibitem[{Zhang et~al.(2022)Zhang, Zhang, Yu, Tang, Tang, Li, and Chen}]{zhang2022subgraph}
Jing Zhang, Xiaokang Zhang, Jifan Yu, Jian Tang, Jie Tang, Cuiping Li, and Hong Chen. 2022.
\newblock \href {https://doi.org/10.18653/v1/2022.acl-long.396} {Subgraph retrieval enhanced model for multi-hop knowledge base question answering}.
\newblock In \emph{Proceedings of the 60th Annual Meeting of the Association for Computational Linguistics (Volume 1: Long Papers)}, pages 5773--5784.

\bibitem[{Zhang et~al.(2023)Zhang, Zhang, Wang, Cao, Huang, Li, Chen, and Li}]{zhang2023fc}
Lingxi Zhang, Jing Zhang, Yanling Wang, Shulin Cao, Xinmei Huang, Cuiping Li, Hong Chen, and Juanzi Li. 2023.
\newblock \href {https://doi.org/10.18653/v1/2023.acl-long.57} {Fc-kbqa: A fine-to-coarse composition framework for knowledge base question answering}.
\newblock In \emph{Proceedings of the 61st Annual Meeting of the Association for Computational Linguistics (Volume 1: Long Papers)}, pages 1002--1017.

\bibitem[{Zhao et~al.(2023)Zhao, Gu, Varma, Luo, Huang, Xu, Wright, Shojanazeri, Ott, Shleifer et~al.}]{zhao2023pytorch}
Yanli Zhao, Andrew Gu, Rohan Varma, Liang Luo, Chien-Chin Huang, Min Xu, Less Wright, Hamid Shojanazeri, Myle Ott, Sam Shleifer, and 1 others. 2023.
\newblock \href {https://doi.org/10.14778/3611540.361156} {Pytorch fsdp: Experiences on scaling fully sharded data parallel}.
\newblock \emph{Proceedings of the VLDB Endowment}, 16(12):3848--3860.

\end{thebibliography}

\appendix

\appendix
%\clearpage

\section{Dataset Details}
\label{appendix:dataset}

The statistics of training and testing sets in the CRAG KG subset is presented in Table~\ref{tab:crag-stats}.

\begin{table}[hbtp]
\caption{Training and testing sets statistics of CRAG.}
\centering
\label{tab:crag-stats}
\begin{tabular}{lll}
\toprule
        & \textbf{Training} & \textbf{Testing} \\
        \midrule
head    & 219      & 187     \\
torso   & 197      & 203     \\
tail    & 188      & 188     \\
\midrule
overall & 604      & 578   \\
\bottomrule
\end{tabular}
\end{table}

The CRAG benchmark includes KGs from open, movie, finance, sports, and music domains. The question types include simple, simple with condition, set, comparison, aggregation, multi-hop, post-processing, and false premise questions. There are also four dynamism considered, including real-time, fast-changing, slow-changing, and static questions. \rev{The proportions of hops required for questions in CRAG are: 43\% requires one hop, 49\% requires two hops, and 8\% requires three hops.}
A detailed statistics about each question type is presented in Table~\ref{tab:crag-types}. 

\begin{table}[hbtp]
\caption{statistics of each question type in different question splits.}
\label{tab:crag-types}
\begin{tabular}{llll}
\toprule
                     & \textbf{head} & \textbf{torso} & \textbf{tail} \\
\midrule
open                 & 48   & 48    & 48   \\
finance              & 125  & 134   & 112  \\
movie                & 121  & 120   & 118  \\
music                & 35   & 41    & 14   \\
sports               & 77   & 57    & 84   \\
\midrule
simple               & 205  & 218   & 187  \\
simple\_w\_condition & 59   & 43    & 57   \\
set                  & 19   & 12    & 13   \\
comparison           & 41   & 44    & 34   \\
aggregation          & 33   & 35    & 38   \\
multi-hop            & 14   & 13    & 11   \\
post-processing      & 6    & 6     & 6    \\
false premise        & 29   & 29    & 30   \\
\midrule
real-time            & 78   & 85    & 72   \\
fast-changing        & 67   & 59    & 76   \\
slow-changing        & 46   & 51    & 45   \\
static               & 215  & 205   & 183 \\
\bottomrule
\end{tabular}
\end{table}

The Open domain question set of the Head2Tail benchmark consists of in total 9219 questions, each of the head, torso, tail data split consist of 3073 questions. We randomly sampled 375 questions from each data split to formulate the testing set, which guarantees a margin of error of less than 5\%. 

The licenses of CRAG and Head2Tail are Attribution-NonCommercial 4.0 International.

The licenses of QALD-10-en, WebQSP, AdvHotpotQA, and CWQ are MIT license, CC-BY-SA-4.0, CC-BY-SA-4.0, and Apache-2.0.

\section{Advantages of KERAG over State-of-the-arts.}
\label{appendix:compare}

\rev{We further discuss the advantages of KERAG over the state-of-the-arts: ToG~\cite{sunthink} and ToG-2~\cite{ma2025thinkongraph}: 1) Our retrieval process is more relax: instead of prompting the LLMs to specify the retrieval paths, we gather all relevant neighborhood and rely more on the strong summarization ability of LLMs to perform filtering and CoT-based summarization. 2) Our method is more efficient: First, our filtering and expansion decisions are made at a schema level whereas the existing approaches make instance-level decisions, so we are much more efficient. Second, assuming $D$ hops, we invoke LLM $D+2$ times, whereas ToG and ToG-2 invoke LLM $ND+D+2$ times ($N$ is \#paths investigated). 3) Our method is more generally applicable. It applies not only to SPARQL-based KGs, but also to API-based KGs, where each entity node may contain long text, with attributes in raw text, JSON style, or tabular format. As such, we can run KERAG on the complex CRAG benchmark, but cannot easily adapt ToG, or ToG-2 on CRAG. }

%\rev{As a result, our approach achieves an overall accuracy uplift of 8.6\% and 2.7\% compared to ToG and ToG-2 as shown in Table~\ref{tab:qald-results} and outperforms StructGPT by 7\% in truthfulness (Table~\ref{tab:baseline_htt}).

\section{Baseline Details}
\label{appendix:baseline}

\begin{itemize}
    \item GPT-4o~\cite{hurst2024gpt}: The Generative Pre-trained Transformers series consists of advanced large language models developed by OpenAI. We chose GPT-4o as the representative for our evaluation. The model is proprietary and can only be accessed via API functions.
    \item GPT-4o (tool): The tool-use variant of the GPT-4o model. We designed a prompt with KG access interfaces to allow the GPT-4o model writes its own API function calls or SPARQL queries. In-context examples are provided to the model. Please check Appendix~\ref{appendix:prompt-tool} for detailed prompts.
    \item Llama-3.1-70B-Instruct~\cite{dubey2024llama}: The Llama-3.1 series is a new collection of large language models released by Meta in July 2024. We adopted the Llama-3.1-70B model with instruct settings to conduct the experiments.
    \item Llama-3.1-70B-Instruct (tool): Similar to GPT-4o (tool), we allow the Llama-3.1-70B-Instruct model to write API function calls or SPARQL queries.
    \item db3~\cite{xia2024winning}: db3 is one of the winning solutions of Meta KDD Cup 2024, which represents the state-of-the-art performance on CRAG benchmark. The design of db3 jointly considers the inputs from KG and web content. We enter ``<EMPTY>'' into the web content module of db3 to adapt it to our KGQA settings.
    \item apex~\cite{ouyang2024revisiting}: apex is one of the winning solutions of Meta KDD Cup 2024, which represents the state-of-the-art performance on CRAG benchmark. A router-based adaptive RAG pipeline is introduced to retrieve relevant KG content for answer generation. Similar to db3, we input ``<EMPTY>'' as the web content input to the model to adapt the apex solution to our KGQA settings.
    \item WikiSP~\cite{xu2023fine}: WikiSP is a semantic parsing KGQA approach. Following the instructions given by WikiSP, we trained the semantic parser on Head2Tail benchmark and replaced the original GPT-3 QA module with the advanced GPT-4o module in the experiments. WikiSP serves as the state-of-the-art representative of semantic parsing KGQA approaches.
    \item StructGPT~\cite{jiang2023structgpt}: StructGPT is a general framework that facilitate the reasoning over structured data. Structured data content such as KGs is retrieved and linearized for the reasoning of LLMs. We selected StructGPT as the state-of-the-art representative for LLM-based KGQA research. Similar to WikiSP, we replace the original ChatGPT core model used by StructGPT with the up-to-date GPT-4o model for fair comparison.
    \item \rev{sparql-qa~\cite{borroto2022sparql}: sparql-qa uses a neural architecture combining NMT and NER with input processing and QQT format to translate natural language to SPARQL.}
    \item \rev{Decaf~\cite{yu2022decaf}: Decaf jointly generates answers and logical forms for knowledge base question answering, combines their advantages, uses text retrieval instead of entity linking to enhance generality.}
    \item \rev{CoK~\cite{lichain}: CoK enhances LLMs by dynamically incorporating heterogeneous knowledge sources through three stage reasoning preparation, dynamic knowledge adapting with an adaptive query generator, and answer consolidation to reduce hallucination and improve factual accuracy.}
    \item \rev{CBR~\cite{das2021case}: CBR is a neuro-symbolic method. It retrieves similar question cases, reuses their logical form components, and revises the generated form via KB embeddings to handle complex KBQA and unseen relations.} 
    \item \rev{ToG~\cite{sunthink}: ToG integrates large language models (LLMs) with knowledge graphs (KGs). Through beam search, LLMs are used to iteratively explore reasoning paths on KGs, so as to enhance the deep reasoning ability of LLMs, improve knowledge traceability and correctness, and achieve state-of-the-art performance on multiple datasets.}
    \item \rev{ToG-2~\cite{ma2025thinkongraph}: Following ToG, ToG-2 is a hybrid RAG framework that tightly couples knowledge graphs and documents for iterative retrieval, enabling deep and faithful reasoning in LLMs with state-of-the-art performance on multiple datasets.}
\end{itemize}

\section{Evaluation Details}
\label{appendix:evaluation}

We adopted the auto-evaluation scheme used in CRAG~\cite{yang2024crag} to evaluate the correctness of the answers on both CRAG and Head2Tail benchmarks. According to~\cite{yang2024crag}, the detailed criterion for accurate, missing, and hallucination is defined as follows:

\begin{itemize}
    \item Correct: The response correctly answers the user’s question and contains no hallucinated content, or the response provides a useful answer to the user’s question but may contain minor errors that do not harm the usefulness of the answer
    \item Missing: The response is “I don’t know”, “I’m sorry I can’t find ...”, a system error such as an empty response, or a request from the system to clarify the original question.
    \item Incorrect: The response provides wrong or irrelevant information to answer the user’s question.
\end{itemize}

We adopted the Llama-3.1-70B-Instruct model as the auto-evaluation critic to evaluate all the responses. The prompts used for auto-evaluation are presented in Appendix~\ref{appendix:prompt-auto}.

\rev{We present the auto-evaluation performance of our Llama-3.1-70B-Instruct model in Table~\ref{tab:auto-eval}.}

\begin{table}[htbp]
\caption{\rev{The Auto-evaluation Performance of Llama-3.1 70B Instruct Model.}}
\label{tab:auto-eval}
\centering
\begin{tabular}{ll}
\toprule
                    & \textbf{Llama-3.1 70B} \\
\midrule
\textbf{correct answers}     & 98.2\%        \\
\textbf{incorrect answers}   & 96.8\%        \\
\textbf{missing}             & 100\%         \\
\textbf{overall reliability} & 98.4\%       \\
\bottomrule
\end{tabular}
\end{table}

\section{Configuration Details}
\label{appendix:config}

All experiments were performed on four NVIDIA A100 GPUs (80GB). 

For the CRAG benchmark, the Llama-3.1-70B-Instruct model is used as the base model for planning. We adopted FSDP~\cite{zhao2023pytorch} and LoRA~\cite{hu2021lora} to efficiently fine-tune the summarizer, which is based on the Llama-3.1-8B-Instruct model. The temperatures are set to 0 for all modules. The summarizer is fine-tuned for 20 epochs (in total 6 GPU hours), with a learning rate of 5e-5. \rev{The proportion of CoT data in the training set is $81.6\%$. The non-CoT data is merged with CoT data for joint training and the training process is one-off.}

For the configurations of KERAG on Head2Tail dataset, Llama-3-8B-Instruct is the base model used for planning, and summarization. Chain-of-Thought reasoning and summarizer finetuning are not applied to the Head2Tail dataset. Temperatures of the models are set to 0 for stable generation of the answer. We include a 3-gram similarity based method to first identify the potential (entity, predicate) pairs (maximum three candidates per query) based on the entity and relation lists in Head2Tail and then apply the Main Entity Extraction prompts to identify the main entity of the query. Since there is only one DBPedia KG in Head2Tail and the questions in Head2Tail are not time-sensitive, we don't need to perform domain and time information extraction. The similarity-based filtering is applied. Specifically, we use DPR~\cite{karpukhin2020dense} to compute and rank the similarity between the candidate predicates in the retrieved triples and the query with the main entity removed. Top-30 predicates are preserved in the retrieval step for the summarizer to generate responses. Specifically, the average input number of predicates is 53.2, and up to 30 predicates are selected for each question.

\rev{The configurations of KERAG on the QALD-10-en, WebQSP, AdvHotpotQA, and CWQ datasets are as follow: We use Llama-3-8B-Instruct as the base model for the planning step.}
The llama-7b-wikiwebquestions-qald7 model\footnote{https://huggingface.co/stanford-oval/llama-7b-wikiwebquestions-qald7}~\cite{xu2023fine} is used to aid the process of main entity extraction. The plain GPT-4o model is used as the summarizer of the model. Chain-of-Thought reasoning and summarizer finetuning are not applied. \rev{Similarity-based DPR filtering is used to filter irrelevant predicates. The planning and summarization prompts are similar to those used on Head2Tail.}

\section{Head-Torso-Tail Experiment Results}
\label{appendix:exp-htt}

We present the detailed results regarding the head/torso/tail categorization of CRAG and Head2Tail datasets in Tables~\ref{tab:baseline_benchmarking_crag_full} and~\ref{tab:baseline_benchmarking_htt_full}.

\begin{table}[t!]
  \caption{Head/torso/tail experimental results on CRAG.}

  \label{tab:baseline_benchmarking_crag_full}
\centering
\scriptsize
\begin{tabular}{llrrrrr}
\toprule &
 \textbf{Model} & \textbf{A} & \textbf{H} & \textbf{M} & \textbf{T} \\
\midrule
head & GPT-4o   &     0.401               &             \textbf{0.086}           &    0.513               &      0.316          \\
& Llama   &   0.390                 &         0.096               &        0.513           &    0.294            \\
\cmidrule{2-6}
& KERAG (8B)   &    0.738           &    0.225             &   0.037      &    0.513        \\
& KERAG   &     \textbf{0.754}               &  0.219                      &   \textbf{0.027}                &     \textbf{0.535}           \\
\midrule
torso & GPT-4o   &      0.355              &                 \textbf{0.089}       &    0.557               &       0.266         \\
& Llama   &      0.305              &     \textbf{0.089}                   &        0.606           &   0.217             \\
\cmidrule{2-6}
& KERAG (8B)   &     0.714          &     0.192            &   0.094      & 0.522            \\
& KERAG   &     \textbf{0.734}               &     0.207                   &      \textbf{0.059}             &     \textbf{0.527}          \\
\midrule
tail & GPT-4o   &        0.266            &    0.096                    &    0.638               &       0.170        \\
& Llama   &    0.223                &          \textbf{0.053}              &       0.723            &   0.170             \\
\cmidrule{2-6}
& KERAG (8B)   &    0.686           &     0.207            &     \textbf{0.106}    &   0.479         \\
& KERAG   &     \textbf{0.707}            &    0.181                    &   0.112              &  \textbf{0.527}             \\
\bottomrule
\end{tabular}
\vspace{-1em}
\end{table}

\begin{table}[t!]
	\caption{Head/torso/tail experimental results on Head2Tail.}

	\label{tab:baseline_benchmarking_htt_full}
	\centering
	\scriptsize
	\begin{tabular}{llrrrrr}
		\toprule &
		\textbf{Model} & \textbf{A} & \textbf{H} & \textbf{M} & \textbf{T} \\
		\midrule
		head & GPT-4o   &      0.613              &        0.181                &      0.205             &  0.432              \\
		& Llama   &   0.629                 &     0.197                   &      0.173             &   0.432             \\
        \cmidrule{2-6}
		& KERAG   &   \textbf{0.883}                 &       \textbf{0.067}                 &     \textbf{0.051}              &    \textbf{0.816}            \\ 
		\midrule
        torso & GPT-4o   &      0.517              &    0.133                    &        0.349           &       0.384         \\
		& Llama   &   0.440                 &   0.139                     &     0.421              &     0.301                     \\
        \cmidrule{2-6}
		& KERAG   &    \textbf{0.917}              &        \textbf{0.045}            &        \textbf{0.037}         &   \textbf{0.872}          \\
		\midrule
		tail & GPT-4o   & 0.376                   &      0.165                  &     0.459              &  0.211              \\
		& Llama   &    0.288                &  0.152                      &       0.560            &    0.136            \\
        \cmidrule{2-6}
		& KERAG   &  \textbf{0.925}                &  \textbf{0.035}                  &      \textbf{0.040}           &   \textbf{0.891}          \\
		\bottomrule
	\end{tabular}
    \vspace{-1em}
\end{table}

\section{Full Ablation Study Results}
\label{appendix:exp-ablation}

The full ablation study results with head/torso/tail categorization on CRAG are presented in Table~\ref{tab:ablation_crag_full}.

\begin{figure}[t!]
  \centering
\includegraphics[width=1.0\linewidth]{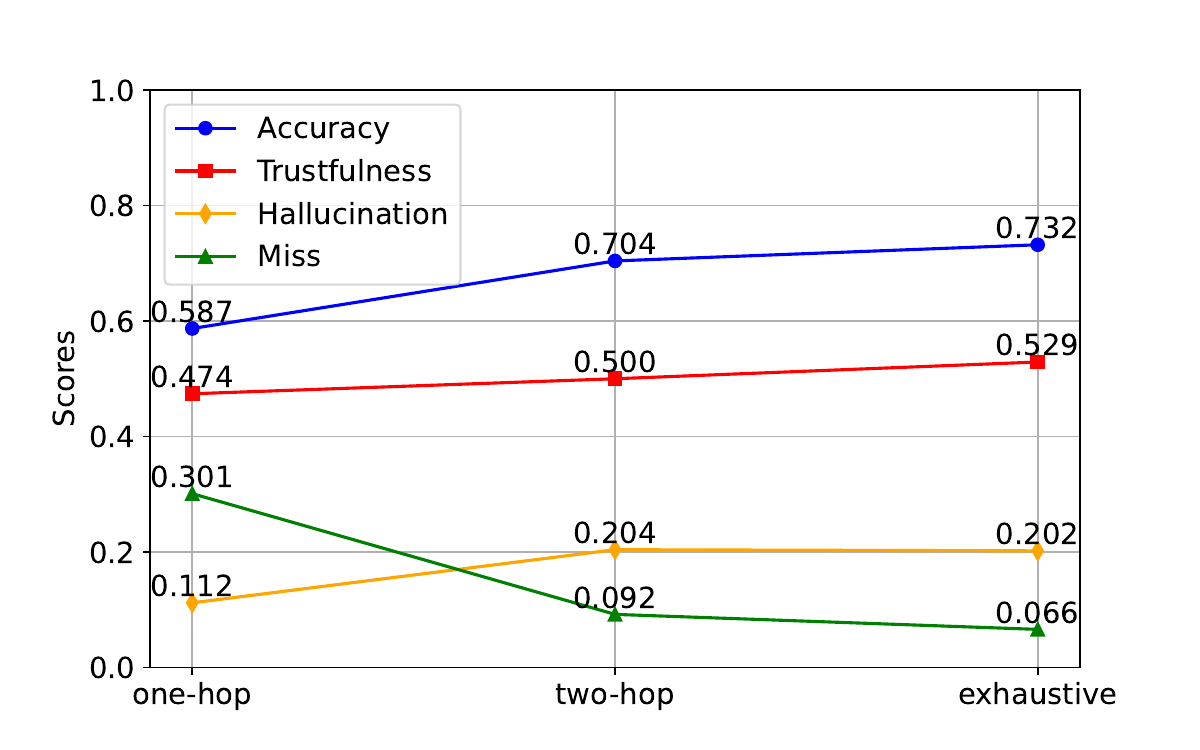}
\vspace{-1em}
  \caption{Comparison between one-hop, two-hop, and exhaustive retrieval on CRAG.}
  \label{fig:discussion}
\vspace{-1em}
\end{figure}

\begin{table}[hbtp]
  \caption{Ablation Study on CRAG.}
  \label{tab:ablation_crag_full}

\centering
\footnotesize
\begin{tabular}{llrrrrr}
\toprule &
 \textbf{Model} & \textbf{A} & \textbf{H} & \textbf{M} & \textbf{T} \\
\midrule
head &   KERAG    &     0.754               &  0.219                      &   0.027                &     0.535           \\
&  - multihop  &     0.626               &        0.150                &   0.225                &         0.476       \\
&  - filter  &     0.738               &                0.246        &  0.016                 &          0.492      \\
&  - finetuning  &      0.663              &          0.203              &  0.134                 &  0.460              \\
&  - CoT  &     0.658               &     0.342                   &    0.000               &      0.316          \\
\midrule
torso & KERAG   &     0.734               &     0.207                   &      0.059             &     0.527          \\
&  - multihop  &       0.596             &                0.094        &     0.310              &        0.502        \\
&  - filter  &     0.704               &             0.246           &   0.049                &        0.458        \\
&  - finetuning  &        0.640            &           0.172             &    0.187               &       0.468         \\
&  - CoT  &     0.640               &          0.360              &               0.000    & 0.281               \\
\midrule
tail & KERAG   &     0.707            &    0.181                    &   0.112              &  0.527             \\
&  - multihop  &      0.537              &            0.096            &   0.367                &          0.441      \\
&  - filter  &      0.703              &            0.202            &     0.074              & 0.501               \\
&  - finetuning  &    0.574                &          0.144              &      0.282             &     0.431           \\
&  - CoT  &   0.649                 &                  0.340      &   0.010                &     0.309           \\
\midrule
overall & KERAG   &     0.732               &                0.202        & 0.066                  & 0.529               \\
&  - multihop  &  0.587                  &            0.112            &       0.301            &       0.474         \\
&  - filter  &    0.721                &                    0.232    &  0.047                 &         0.490       \\
&  - finetuning  &       0.626             &    0.173                    &                 0.201  &       0.453         \\
&  - CoT  &    0.649                &                  0.348      &  0.003                 &         0.301       \\
\bottomrule
\end{tabular}
\vspace{-1em}
\end{table}

\section{Templates}
\label{appendix:template}
\begin{block}{Planning Prompts Skeleton}
%\textit{\textbf{Main Entity Extraction}}\\
%\noindent\rule{\linewidth}{0.4pt}
You are given a Query [$Q$], please extract the topic entity [$E$] from the Query.

Determine the domain the query is about. The domain should be one of the following: [Set of KG domains].

[Examples] (Optional)

[Further Instructions] (Optional)

\end{block}

\begin{block}{Filtering Prompts Skeleton}

Given the user query [$Q$] with detected topic entity [$E$] and the current [retrieval plan], you will also have access to the following functions [a set of functions/predicates with ontology] to perform additional hops based on the current plan.

Decide which functions/predicates are relevant to answer the query based on the ontology and if the composed retrieval plan sufficient for the question answering.

[Examples] (Optional)

[Further Instructions] (Optional)

\end{block}

\begin{block}{Summarization Prompts Skeleton}

Please provide a brief answer to the question [$Q$] based on your knowledge and the following content. Answer ``I don't know'' if you are not confident of your answer. Please think step by step.

[Retrieved Content (Triple form)]

\end{block}

\section{Prompts}
\label{appendix:prompt}

We present the detailed prompts used in our experiments. Italicized text enclosed in pointed brackets represents the input variables.

\subsection{Planning Prompts}
\label{appendix:prompt-parser}

\begin{block}{Planning Prompts of CRAG}
\textit{\textbf{Main Entity and Domain Extraction}}\\
\noindent\rule{\linewidth}{0.4pt}

You are an agent only outputs JSON. You are given a Query and Query Time. Do the following: 

1) Determine the domain the query is about. The domain should be one of the following: "finance", "sports", "music", "movie", "encyclopedia". If none of the domain applies, use "other". Use "domain" as the key in the result json. 

2) Extract structured information from the query. Include different keys into the result json depending on the domains, amd put them DIRECTLY in the result json. Here are the rules:

For `encyclopedia` and `other` queries, these are possible keys:

-  `main\_entity`: extract the main entity of the query. 

For `finance` queries, these are possible keys:

- `market\_identifier`: stock identifiers including individual company names, stock symbols.

- `metric`: financial metrics that the query is asking about. This must be one of the following: `price`, `dividend`, `P/E ratio`, `EPS`, `marketCap`, and `other`.

For `movie` queries, these are possible keys:

- `movie\_name`: name of the movie

- `movie\_aspect`: if the query is about a movie, which movie aspect the query asks. This must be one of the following: `budget`, `genres`, `original\_language`, `original\_title`, `release\_date`, `revenue`, `title`, `cast`, `crew`, `rating`, `length`.

- `person`: person name related to moves

- `person\_aspect`: if the query is about a person, which person aspect the query asks. This must be one of the following: `acted\_movies`, `directed\_movies`, `oscar\_awards`, `birthday`.

- `year`: if the query is about movies released in a specific year, extract the year

For `music` queries, these are possible keys:
- `artist\_name`: name of the artist

- `artist\_aspect`: if the query is about an artist, extract the aspect of the artist. This must be one of the following: `member`, `birth place`, `birth date`, `lifespan`, `artist work`, `grammy award count`, `grammy award date`.

- `song\_name`: name of the song

- `song\_aspect`: if the query is about a song, extract the aspect of the song. This must be one of the following: `auther`, `grammy award count`, `release country`, `release date`.

For `sports` queries, these are possible keys:
- `sport\_type`: one of `basketball`, `soccer`, `other`

- `tournament`: such as NBA, World Cup, Olympic.

- `team`: teams that user interested in.

Return the results in a FLAT json. 

*NEVER include ANY EXPLANATION or NOTE in the output, ONLY OUTPUT JSON!!!*  

Here are some examples:

\#\#\# Query

what was the volume of trades for rcm on the last day?

Your extracted JSON should be: {"domain": "finance", "market\_identifier": "rcm", "metric": "volume of trades"}

\#\#\# Query

on 2022-10-11, how many points did bulls put up in their game?

Your extracted JSON should be: {"domain": "sports", "sport\_type": "basketball", "team": "chicago bulls"}
\\

\textit{\textbf{CoT-based Time Extraction}}\\
\noindent\rule{\linewidth}{0.4pt}
Please extract the time frame that user interested in. When datetime is not explicitly mentioned, use `Query Time` as default. Enclose your datetime extracted with <>! Use '$\sim$' to indicate time period. 

Here are some examples:

\#\#\# Query

what was the volume of trades for rcm on the last day?

\#\#\# Query Time

02/28/2024, 08:04:08 PT

Question: What is the user interested time frame of the Query? Please think step by step.

Your output: Since the query is asking about the last day of the Query Time, the time frame should be <02/27/2024>.

\#\#\# Query

which team did boston celtics take on in their matchup on 2023-05-29?

\#\#\# Query Time

03/15/2024, 16:05:17 PT

Question: What is the user interested time frame of the Query? Please think step by step.

Your output: The time frame should be <05/29/2023>, which is explicitly stated in the Query.

\#\#\# Query

on which date did sgml distribute dividends the first time?

\#\#\# Query Time

02/28/2024, 08:25:10 PT

Question: What is the user interested time frame of the Query? Please think step by step.

Your output: The time frame should be <02/28/2024>. Since there is no datetime explicitly mentioned and we take the Query Time as the default answer.

\#\#\# Query

what's the schedule looking like for west ham's next game in eng-premier league?

\#\#\# Query Time

03/15/2024, 15:48:32 PT

Question: What is the user interested time frame of the Query? Please think step by step.

Your output: The time frame should be <03/15/2024 $\sim$ future>. Since there is no datetime that can be confidently inferred from the query, we take the Query Time to ($\sim$) future time period as the answer.

\#\#\# Query

on average, what was the daily high stock price of xpev over the past week?

\#\#\# Query Time

02/28/2024, 08:17:40 PT

Question: What is the user interested time frame of the Query? Please think step by step.

Your output: The time frame should be <02/21/2024 $\sim$ 02/28/2024>. Since the question is asking for the value over the past week. We take 02/21/2024 $\sim$ 02/28/2024 time period as the answer. 
\end{block}

\begin{block}{Planning Prompts of Head2Tail}
\textit{\textbf{Main Entity Extraction}}\\
\noindent\rule{\linewidth}{0.4pt}

You are a professional semantic parsing expert. Please provide your answer by giving a brief (entity, predicate) pair. If no appropriate (entity, predicate) pair can be extracted, please respond (No, No).

[Instruction]: The task is to determine the most appropriate (entity, predicate) pair based on a given question and a list of potential (entity, predicate) pair list extracted by a fuzzy-matching-based semantic parser. MUST NOT provide answer to the question directly. 

[EXAMPLE]: Question: What aircraft bomber was used by the South Vietnam Air Force? Pair List: [('Vietnam', 'Air\_Force'), ('South\_Vietnam\_Air\_Force', 'aircraftBomber')]; Your Answer: ('South\_Vietnam\_Air\_Force', 'aircraftBomber')

[TASK]: Select the most appropriate (entity, predicate) based on this question and entity pair list: Question: <\textit{Input Question}> Pair List: <\textit{Candidate List}>.
\end{block}

\subsection{Filtering Prompts}
\label{appendix:prompt-filterer}

We present an exemplar filtering prompt used in the finance KG of CRAG as an example. The filtering prompts on other KGs can be generated by replacing the function descriptions and in context example accordingly. 

\begin{block}{Filtering Prompts of CRAG}
\textit{\textbf{Tool-based Filtering on Finance KG}}\\
\noindent\rule{\linewidth}{0.4pt}
You have access to the following functions:

Use the function 'finance\_get\_price\_history' to 'Return daily Open price, Close price, High price, Low price and trading Volume.':
\{"type": "function", "function": \{"name": "finance\_get\_price\_history", "description": "Return daily Open price, Close price, High price, Low price and trading Volume.", "parameters": \{"type": "object", "properties": \{"ticker\_name": \{"type": "string", "description": "The ticker name of the stock interested."\}\}, "required": ["ticker\_name"]\}\}\}

Use the function 'finance\_get\_detailed\_price\_history' to 'Return minute-level Open price, Close price, High price, Low price and trading Volume.':
\{"type": "function", "function": \{"name": "finance\_get\_detailed\_price\_history", "description": "Return minute-level Open price, Close price, High price, Low price and trading Volume.", "parameters": \{"type": "object", "properties": \{"ticker\_name": \{"type": "string", "description": "The ticker name of the stock interested."\}\}, "required": ["ticker\_name"]\}\}\}

Use the function 'finance\_get\_dividends\_history' to 'Return dividend history of a ticker.':
\{"type": "function", "function": \{"name": "finance\_get\_dividends\_history", "description": "Return dividend history of a ticker.", "parameters": \{"type": "object", "properties": \{"ticker\_name": \{"type": "string", "description": "The ticker name of the stock interested."\}\}, "required": ["ticker\_name"]\}\}\}

Use the function 'finance\_get\_market\_capitalization' to 'Return the market capitalization of a ticker.':
\{"type": "function", "function": \{"name": "finance\_get\_market\_capitalization", "description": "Return the market capitalization of a ticker.", "parameters": \{"type": "object", "properties": \{"ticker\_name": \{"type": "string", "description": "The ticker name of the stock interested."\}\}, "required": ["ticker\_name"]\}\}\}

Use the function 'finance\_get\_eps' to 'Return earnings per share of a ticker.':
\{"type": "function", "function": \{"name": "finance\_get\_eps", "description": "Return earnings per share of a ticker.", "parameters": \{"type": "object", "properties": \{"ticker\_name": \{"type": "string", "description": "The ticker name of the stock interested."\}\}, "required": ["ticker\_name"]\}\}\}

Use the function 'finance\_get\_pe\_ratio' to 'Return price-to-earnings ratio of a ticker.':
\{"type": "function", "function": \{"name": "finance\_get\_pe\_ratio", "description": "Return price-to-earnings ratio of a ticker.", "parameters": \{"type": "object", "properties": \{"ticker\_name": \{"type": "string", "description": "The ticker name of the stock interested."\}\}, "required": ["ticker\_name"]\}\}\}

Use the function 'finance\_get\_info' to 'Return rough meta data of a ticker.':
\{"type": "function", "function": \{"name": "finance\_get\_info", "description": "Return rough meta data of a ticker.", "parameters": \{"type": "object", "properties": \{"ticker\_name": \{"type": "string", "description": "The ticker name of the stock interested."\}\}, "required": ["ticker\_name"]\}\}\}

    If you choose to call a function ONLY reply in the following format with no prefix or suffix:

    Question: What is the price of Meta currently?
    Your answer: <function=detailedPriceHistoryTool></function>

    Reminder:

    - Function calls MUST follow the specified format, start with <function= and end with </function>
    
    - Only call one function at a time
    
    - Put the entire function call reply on one line
    
    - If there is no function call available, answer I don't know.'
    
Question: <\textit{Question}>
\end{block}

\begin{block}{Multi-hop Prompts of Head2Tail}
\textit{\textbf{Multi-hop Retrieval Boundary Prompts}}\\
\noindent\rule{\linewidth}{0.4pt}

You will be given a QUESTION and a set of retrieved TRIPLEs from DBpedia. Your task is to indicate whether the currently retrieved content is sufficient for you to answer the QUESTION. If you need to have more retrieved triples, respond <NO>. If you think the subject entity is wrongly linked, respond <NA>. Otherwise, if you think the current information is sufficient, respond <YES>. Only answer <NO>/<NA>/<YES>!!!

QUESTION: <\textit{Input Question}>

DBpedia TRIPLEs: <\textit{Retrieved triples}>
\end{block}

\subsection{Summarization Prompts}
\label{appendix:prompt-summarizer}

\begin{block}{Summarization Prompts of CRAG}
\textit{\textbf{CoT-based Prompt}}\\
\noindent\rule{\linewidth}{0.4pt}
Please provide a brief answer as short as possible to the question based on your own knowledge and the following relevant CONTENT extracted from Knowledge Base. Answer "I don't know" if you are not confident of your answer. Please think step by step.

The current query time is: <\textit{Query Time}>.

<\textit{Question}>

CONTENT: <\textit{Refined KG Content}>
\\

\textit{\textbf{Base Prompt}}\\
\noindent\rule{\linewidth}{0.4pt}
Please provide a brief answer as short as possible to the question based on your own knowledge and the following relevant CONTENT extracted from Knowledge Base. Answer "I don't know" if you are not confident of your answer. 

The current query time is: <\textit{Query Time}>.

<\textit{Question}>

CONTENT: <\textit{Refined KG Content}>

\end{block}

\begin{block}{Summarization Prompts of Head2Tail}
\textit{\textbf{Summarization Prompt}}\\
\noindent\rule{\linewidth}{0.4pt}
Please provide a brief answer as short as possible to the question based on your own knowledge and the following relevant TRIPLEs (subject, predicate, object) from DBpedia. Answer "I don't know" if you are not confident of your answer.

TRIPLE: <\textit{triple\_1}>

TRIPLE: <\textit{triple\_2}>

......

TRIPLE: <\textit{triple\_n}>

<\textit{Question}>
\end{block}

\subsection{Baselines Tool-use Prompts}
\label{appendix:prompt-tool}

\begin{block}{Tool-use Prompts for GPT-4o and Llama}
\textit{\textbf{Tool-use Prompts of CRAG}}\\
\noindent\rule{\linewidth}{0.4pt}

Please select the proper APIs to retrieve relevant content to the question.

You have access to the following functions:

Use the function 'open\_search\_entity\_by\_name' to 'Search for entities by name in the Open domain.':
\{"type": "function", "function": \{"name": "open\_search\_entity\_by\_name", "description": "Search for entities by name in the Open domain.", "parameters": \{"type": "object", "properties": \{"query": \{"type": "string", "description": "The name of the entity interested."\}\}, "required": ["query"]\}\}\}

Use the function 'open\_get\_entity' to 'Retrieve detailed information about an entity in the Open domain.':
\{"type": "function", "function": \{"name": "open\_get\_entity", "description": "Retrieve detailed information about an entity in the Open domain.", "parameters": \{"type": "object", "properties": \{"entity": \{"type": "string", "description": "The name of the entity interested."\}\}, "required": ["entity"]\}\}\}

Use the function 'movie\_get\_person\_info' to 'Get information about a person related to movies.':
\{"type": "function", "function": \{"name": "movie\_get\_person\_info", "description": "Get information about a person related to movies.", "parameters": \{"type": "object", "properties": \{"person\_name": \{"type": "string", "description": "The name of the person interested."\}\}, "required": ["person\_name"]\}\}\}

Use the function 'movie\_get\_movie\_info' to 'Get information about a movie.':
\{"type": "function", "function": \{"name": "movie\_get\_movie\_info", "description": "Get information about a movie.", "parameters": \{"type": "object", "properties": \{"movie\_name": \{"type": "string", "description": "The name of the movie interested."\}\}, "required": ["movie\_name"]\}\}\}

Use the function 'movie\_get\_year\_info' to 'Get information about movies released in a specific year.':
\{"type": "function", "function": \{"name": "movie\_get\_year\_info", "description": "Get information about movies released in a specific year.", "parameters": \{"type": "object", "properties": \{"year": \{"type": "string", "description": "The year interested."\}\}, "required": ["year"]\}\}\}

Use the function 'movie\_get\_person\_info\_by\_id' to 'Get person information by their unique ID.':
\{"type": "function", "function": \{"name": "movie\_get\_person\_info\_by\_id", "description": "Get person information by their unique ID.", "parameters": \{"type": "object", "properties": \{"person\_id": \{"type": "integer", "description": "The person id interested."\}\}, "required": ["person\_id"]\}\}\}

Use the function 'movie\_get\_movie\_info\_by\_id' to 'Get movie information by its unique ID.':
\{"type": "function", "function": \{"name": "movie\_get\_movie\_info\_by\_id", "description": "Get movie information by its unique ID.", "parameters": \{"type": "object", "properties": \{"movie\_id": \{"type": "integer", "description": "The movie id interested."\}\}, "required": ["movie\_id"]\}\}\}

Use the function 'finance\_get\_company\_name' to 'Search for company names in the finance domain.':
\{"type": "function", "function": \{"name": "finance\_get\_company\_name", "description": "Search for company names in the finance domain.", "parameters": \{"type": "object", "properties": \{"query": \{"type": "string", "description": "The query interested."\}\}, "required": ["query"]\}\}\}

Use the function 'finance\_get\_ticker\_by\_name' to 'Retrieve the ticker symbol for a given company name.':
\{"type": "function", "function": \{"name": "finance\_get\_ticker\_by\_name", "description": "Retrieve the ticker symbol for a given company name.", "parameters": \{"type": "object", "properties": \{"query": \{"type": "string", "description": "The query interested."\}\}, "required": ["query"]\}\}\}

Use the function 'finance\_get\_price\_history' to 'Return daily Open price, Close price, High price, Low price and trading Volume.':
\{"type": "function", "function": \{"name": "finance\_get\_price\_history", "description": "Return daily Open price, Close price, High price, Low price and trading Volume.", "parameters": \{"type": "object", "properties": \{"ticker\_name": \{"type": "string", "description": "The ticker name of the stock interested."\}\}, "required": ["ticker\_name"]\}\}\}

Use the function 'finance\_get\_detailed\_price\_history' to 'Return minute-level Open price, Close price, High price, Low price and trading Volume.':
\{"type": "function", "function": \{"name": "finance\_get\_detailed\_price\_history", "description": "Return minute-level Open price, Close price, High price, Low price and trading Volume.", "parameters": \{"type": "object", "properties": \{"ticker\_name": \{"type": "string", "description": "The ticker name of the stock interested."\}\}, "required": ["ticker\_name"]\}\}\}

Use the function 'finance\_get\_dividends\_history' to 'Return dividend history of a ticker.':
\{"type": "function", "function": \{"name": "finance\_get\_dividends\_history", "description": "Return dividend history of a ticker.", "parameters": \{"type": "object", "properties": \{"ticker\_name": \{"type": "string", "description": "The ticker name of the stock interested."\}\}, "required": ["ticker\_name"]\}\}\}

Use the function 'finance\_get\_market\_capitalization' to 'Return the market capitalization of a ticker.':
\{"type": "function", "function": \{"name": "finance\_get\_market\_capitalization", "description": "Return the market capitalization of a ticker.", "parameters": \{"type": "object", "properties": \{"ticker\_name": \{"type": "string", "description": "The ticker name of the stock interested."\}\}, "required": ["ticker\_name"]\}\}\}

Use the function 'finance\_get\_eps' to 'Return earnings per share of a ticker.':
\{"type": "function", "function": \{"name": "finance\_get\_eps", "description": "Return earnings per share of a ticker.", "parameters": \{"type": "object", "properties": \{"ticker\_name": \{"type": "string", "description": "The ticker name of the stock interested."\}\}, "required": ["ticker\_name"]\}\}\}

Use the function 'finance\_get\_pe\_ratio' to 'Return price-to-earnings ratio of a ticker.':
\{"type": "function", "function": \{"name": "finance\_get\_pe\_ratio", "description": "Return price-to-earnings ratio of a ticker.", "parameters": \{"type": "object", "properties": \{"ticker\_name": \{"type": "string", "description": "The ticker name of the stock interested."\}\}, "required": ["ticker\_name"]\}\}\}

Use the function 'finance\_get\_info' to 'Return rough meta data of a ticker.':
\{"type": "function", "function": \{"name": "finance\_get\_info", "description": "Return rough meta data of a ticker.", "parameters": \{"type": "object", "properties": \{"ticker\_name": \{"type": "string", "description": "The ticker name of the stock interested."\}\}, "required": ["ticker\_name"]\}\}\}

Use the function 'music\_search\_artist\_entity\_by\_name' to 'Search for music artists by name.':
\{"type": "function", "function": \{"name": "music\_search\_artist\_entity\_by\_name", "description": "Search for music artists by name.", "parameters": \{"type": "object", "properties": \{"artist\_name": \{"type": "string", "description": "The name of the artist interested."\}\}, "required": ["artist\_name"]\}\}\}

Use the function 'music\_search\_song\_entity\_by\_name' to 'Search for songs by name.':
\{"type": "function", "function": \{"name": "music\_search\_song\_entity\_by\_name", "description": "Search for songs by name.", "parameters": \{"type": "object", "properties": \{"song\_name": \{"type": "string", "description": "The name of the song interested."\}\}, "required": ["song\_name"]\}\}\}

Use the function 'music\_get\_billboard\_attributes' to 'Get attributes of a song from Billboard rankings.':
\{"type": "function", "function": \{"name": "music\_get\_billboard\_attributes", "description": "Get attributes of a song from Billboard rankings.", "parameters": \{"type": "object", "properties": \{"date": \{"type": "string", "description": "The date you are interested."\}, "attribute": \{"type": "string", "description": "The attribute you are interested."\}, "song\_name": \{"type": "string", "description": "The song you are interested."\}\}, "required": ["date", "attribute", "song\_name"]\}\}\}

Use the function 'music\_get\_billboard\_rank\_date' to 'Get Billboard ranking for a specific rank and date.':
\{"type": "function", "function": \{"name": "music\_get\_billboard\_rank\_date", "description": "Get Billboard ranking for a specific rank and date.", "parameters": \{"type": "object", "properties": \{"rank": \{"type": "integer", "description": "The rank you are interested."\}, "date": \{"type": "string", "description": "The date you are interested."\}\}, "required": ["rank"]\}\}\}

Use the function 'music\_grammy\_get\_best\_album\_by\_year' to 'Get the Grammy Album of the Year for a specific year.':
\{"type": "function", "function": \{"name": "music\_grammy\_get\_best\_album\_by\_year", "description": "Get the Grammy Album of the Year for a specific year.", "parameters": \{"type": "object", "properties": \{"year": \{"type": "integer", "description": "The year interested."\}\}, "required": ["year"]\}\}\}

Use the function 'music\_grammy\_get\_all\_awarded\_artists' to 'Get all artists awarded the Grammy Best New Artist.':
\{"type": "function", "function": \{"name": "music\_grammy\_get\_all\_awarded\_artists", "description": "Get all artists awarded the Grammy Best New Artist.", "parameters": \{"type": "object", "properties": \{\}, "required": []\}\}\}

Use the function 'music\_grammy\_get\_award\_count\_by\_song' to 'Get the total Grammy awards won by a song.':
\{"type": "function", "function": \{"name": "music\_grammy\_get\_award\_count\_by\_song", "description": "Get the total Grammy awards won by a song.", "parameters": \{"type": "object", "properties": \{"song\_name": \{"type": "string", "description": "The song interested."\}\}, "required": ["song\_name"]\}\}\}

Use the function 'music\_grammy\_get\_best\_song\_by\_year' to 'Get the Grammy Song of the Year for a specific year.':
\{"type": "function", "function": \{"name": "music\_grammy\_get\_best\_song\_by\_year", "description": "Get the Grammy Song of the Year for a specific year.", "parameters": \{"type": "object", "properties": \{"year": \{"type": "integer", "description": "The year interested."\}\}, "required": ["year"]\}\}\}

Use the function 'music\_grammy\_get\_award\_date\_by\_artist' to 'Get the years an artist won a Grammy award.':
\{"type": "function", "function": \{"name": "music\_grammy\_get\_award\_date\_by\_artist", "description": "Get the years an artist won a Grammy award.", "parameters": \{"type": "object", "properties": \{"artist\_name": \{"type": "string", "description": "The artist interested."\}\}, "required": ["artist\_name"]\}\}\}

Use the function 'music\_grammy\_get\_best\_album\_by\_year' to 'Get the Grammy Album of the Year for a specific year.':
\{"type": "function", "function": \{"name": "music\_grammy\_get\_best\_album\_by\_year", "description": "Get the Grammy Album of the Year for a specific year.", "parameters": \{"type": "object", "properties": \{"year": \{"type": "integer", "description": "The year interested."\}\}, "required": ["year"]\}\}\}

Use the function 'music\_grammy\_get\_all\_awarded\_artists' to 'Get all artists awarded the Grammy Best New Artist.':
\{"type": "function", "function": \{"name": "music\_grammy\_get\_all\_awarded\_artists", "description": "Get all artists awarded the Grammy Best New Artist.", "parameters": \{"type": "object", "properties": \{\}, "required": []\}\}\}

Use the function 'music\_get\_artist\_birth\_place' to 'Return the birth place country code (2-digit) for the input artist.':
\{"type": "function", "function": \{"name": "music\_get\_artist\_birth\_place", "description": "Return the birth place country code (2-digit) for the input artist.", "parameters": \{"type": "object", "properties": \{"artist\_name": \{"type": "string", "description": "The name of the artist interested."\}\}, "required": ["artist\_name"]\}\}\}

Use the function 'music\_get\_artist\_birth\_date' to 'Return the birth date of the artist.':
\{"type": "function", "function": \{"name": "music\_get\_artist\_birth\_date", "description": "Return the birth date of the artist.", "parameters": \{"type": "object", "properties": \{"artist\_name": \{"type": "string", "description": "The name of the artist interested."\}\}, "required": ["artist\_name"]\}\}\}

Use the function 'music\_get\_members' to 'Return the member list of a band / person.':
\{"type": "function", "function": \{"name": "music\_get\_members", "description": "Return the member list of a band / person.", "parameters": \{"type": "object", "properties": \{"band\_name": \{"type": "string", "description": "The name of the band / person interested."\}\}, "required": ["band\_name"]\}\}\}

Use the function 'music\_get\_lifespan' to 'Return the lifespan of the artist.':
\{"type": "function", "function": \{"name": "music\_get\_lifespan", "description": "Return the lifespan of the artist.", "parameters": \{"type": "object", "properties": \{"artist\_name": \{"type": "string", "description": "The name of the artist interested."\}\}, "required": ["artist\_name"]\}\}\}

Use the function 'music\_get\_song\_author' to 'Get the author of a song.':
\{"type": "function", "function": \{"name": "music\_get\_song\_author", "description": "Get the author of a song.", "parameters": \{"type": "object", "properties": \{"song\_name": \{"type": "string", "description": "The name of the song interested."\}\}, "required": ["song\_name"]\}\}\}

Use the function 'music\_get\_song\_release\_country' to 'Get the release country of a song.':
\{"type": "function", "function": \{"name": "music\_get\_song\_release\_country", "description": "Get the release country of a song.", "parameters": \{"type": "object", "properties": \{"song\_name": \{"type": "string", "description": "The name of the song interested."\}\}, "required": ["song\_name"]\}\}\}

Use the function 'music\_get\_song\_release\_date' to 'Get the release date of a song.':
\{"type": "function", "function": \{"name": "music\_get\_song\_release\_date", "description": "Get the release date of a song.", "parameters": \{"type": "object", "properties": \{"song\_name": \{"type": "string", "description": "The name of the song interested."\}\}, "required": ["song\_name"]\}\}\}

Use the function 'music\_get\_artist\_all\_works' to 'Return the list of all works of a certain artist.':
\{"type": "function", "function": \{"name": "music\_get\_artist\_all\_works", "description": "Return the list of all works of a certain artist.", "parameters": \{"type": "object", "properties": \{"artist\_name": \{"type": "string", "description": "The name of the artist interested."\}\}, "required": ["artist\_name"]\}\}\}

Use the function 'sports\_soccer\_get\_games\_on\_date' to 'Get soccer games on a specific date. Result includes game attributes such as date, time, GF: GF: Goals For - the number of goals scored by the team in question during the match, GA: Goals Against - the number of goals conceded by the team during the match, xG: Expected Goals - a statistical measure that estimates the number of goals a team should have scored based on the quality of chances they created, xGA: Expected Goals Against - a measure estimating the number of goals a team should have conceded based on the quality of chances allowed to the opponent, Poss: Possession - the percentage of the match time during which the team had possession of the ball.':
\{"type": "function", "function": \{"name": "sports\_soccer\_get\_games\_on\_date", "description": "Get soccer games on a specific date. Result includes game attributes such as date, time, GF: GF: Goals For - the number of goals scored by the team in question during the match, GA: Goals Against - the number of goals conceded by the team during the match, xG: Expected Goals - a statistical measure that estimates the number of goals a team should have scored based on the quality of chances they created, xGA: Expected Goals Against - a measure estimating the number of goals a team should have conceded based on the quality of chances allowed to the opponent, Poss: Possession - the percentage of the match time during which the team had possession of the ball.", "parameters": \{"type": "object", "properties": \{"date": \{"type": "string", "description": "The date interested."\}, "team\_name": \{"type": "string", "description": "The team interested."\}\}, "required": ["date", "team\_name"]\}\}\}

Use the function 'sports\_nba\_get\_games\_on\_date' to 'Get NBA games on a specific date. Result includes game attributes such as team\_name\_home: The full name of the home team, wl\_home: The outcome of the game for the home team, pts\_home: The total number of points scored by the home team.':
\{"type": "function", "function": \{"name": "sports\_nba\_get\_games\_on\_date", "description": "Get NBA games on a specific date. Result includes game attributes such as team\_name\_home: The full name of the home team, wl\_home: The outcome of the game for the home team, pts\_home: The total number of points scored by the home team.", "parameters": \{"type": "object", "properties": \{"date": \{"type": "string", "description": "The date interested."\}, "team\_name": \{"type": "string", "description": "The team interested."\}\}, "required": ["date", "team\_name"]\}\}\}

Use the function 'sports\_nba\_get\_play\_by\_play\_data\_by\_game\_ids' to 'Get NBA play by play data for a set of game ids. Result includes play-by-play event time, description, player etc.':
\{"type": "function", "function": \{"name": "sports\_nba\_get\_play\_by\_play\_data\_by\_game\_ids", "description": "Get NBA play by play data for a set of game ids. Result includes play-by-play event time, description, player etc.", "parameters": \{"type": "object", "properties": \{"game\_ids": \{"type": "list", "description": "List of game ids."\}\}, "required": ["game\_ids"]\}\}\}

    If you choose to call a function ONLY reply in the following format with no prefix or suffix, if you think you need to chain multiple function calls, reply the function that you want to call at the current round and a special token <CONTINUE>:

    Question: What is the price of ABCD currently?
    Your answer: <function=finance\_get\_detailed\_price\_history>\{"ticker\_name": "ABCD"\}</function>

    Question: Can you tell me who directed the movie ABCD?
    Your answer: <function=movie\_get\_movie\_info>\{"movie\_name": "ABCD"\}</function>
    
    Question: Who are the release date of the last song of band ABCD?
    Your answer: <function=music\_get\_artist\_all\_works>\{"band\_name": "ABCD"\}</function>; <CONTINUE>
    
    Question: On 2022-11-11, how many points did Pacers put up in their game?
    Your answer: <function=sports\_nba\_get\_games\_on\_date>\{"date": "2022-11-11", "team\_name": "Indiana Pacers"\}</function>
    
    Question: What is the population of China?
    Your answer: <function=open\_get\_entity>\{"entity": "China"\}</function>

    Question: What is the movie that has id 123456?
    Your answer: <function=movie\_get\_movie\_info>\{"movie\_id": 123456\}</function>

    Reminder:
    
    - Function calls MUST follow the specified format, start with <function= and end with </function>
    
    - Put the entire function call reply on one line
    
    - If there is no function call available, answer I don't know.'
    
    - Reply <CONTINUE> after the function call you choose, seperate them with ;.
    
The current query time is: <\textit{Query Time}> 

<\textit{Question}>\\

\textit{\textbf{Tool-use Prompts of Head2Tail}}\\
\noindent\rule{\linewidth}{0.4pt}

You need to write SPARQL query to access DBPedia to get the answer to the QUESTION. Here is the instruction: 1. Please select the main entity from the QUESTION; 2. Select the proper predicate from the given PREDICATE LIST; 3. Based on the main entity, predicate selected, write a SPARQL query to answer the QUESTION. Enclose the main entity selected with <>; Enclose the predicate selected with []. Only respond the main entity and predicate!!!"

QUESTION: <\textit{Question}>

PREDICATE LIST: <\textit{List of predicate}>

\end{block}

\subsection{Auto-evaluation Prompts}
\label{appendix:prompt-auto}

\begin{block}{Auto-evaluation Prompts}
\textit{\textbf{Auto-evaluation of CRAG}}\\
\noindent\rule{\linewidth}{0.4pt}

Assume you are a human expert in grading predictions given by a model. You are given a question and a model prediction. Judge if the prediction matches the ground truth answer by following these steps:
1: Take it as granted that the Ground Truth is always correct.

2: If the Prediction indicates it is not sure about the answer, "score" should be "0"; otherwise, go the next step.

3: If the Prediction exactly matches the Ground Truth, "score" is 1.

4: If the Prediction does not exactly match the Ground Truth, go through the following steps and likely give a score as 0.

5: If the Ground Truth is a number, "score" is 1 if and only if the Prediction gives a number that almost exactly matches the ground truth.

6: If the Prediction is self-contradictory, "score" must be 0.

7: If the prediction is not answering the question, "score" must be 0.

8: If the prediction is a concise and correct summary of the ground truth, "score" is 1.

9: If ground truth contains a set of items, prediction must contain exactly same items for the score to be 1.

10: Otherwise, "score" is 0.

\#\#\# Output a JSON blob with an "explanation" field explaining your answer as short as possible and an "score" field with value 1 or 0.
You should make the judgment based on provided examples.
Examples:
Question: "which company has higher eps, btu or cma?"
Ground Truth: "cma"
Prediction: "it is not possible to determine which company has a higher eps."
Output: \{"score": 0, "explanation": "The prediction is not sure about the answer."\}

Question: "who is taller, a or b?"
Ground Truth: "a"
Prediction: "The answer is a. a is of height 1.75 m and b is of height 1.82 m. So b is taller."
Output: \{"score": 0, "explanation": "The prediction is self-contradictory."\}

Question: "who among patricia place and lana clarkson is younger?"
Ground Truth: "lana clarkson"
Prediction: "lana clarkson (born april 5, 1962) is younger than patricia clarkson (born december 29, 1959)."
Output: \{"score": 1, "explanation": "The prediction gives the correct answer."\}

Question: "did lakers win more games than warriors in 2024-4?"
Ground Truth: "yes"
Prediction: "yes, lakers won more games than warriors in 2023-2024 season."
Output: \{"score": 0, "explanation": "The prediction is not answering the question."\}

Question: "what is the height of the building where citigroup is headquartered?"
Ground Truth: "151 m"
Prediction: "the height is 915 feet (279 m)."
Output: \{"score": 0, "explanation": "The prediction, 151 m, does not match the ground truth, 279 m."\}

Question: "who were the last 3 prime ministers of uk?"
Ground Truth: "rishi sunak, liz truss, and boris johnson"
Prediction: "1. liz truss, 2. boris johnson, 3.theresa may"
Output: \{"score": 0, "explanation": "The prediction does not contain item, rishi sunak, that is in the grount truth."\}

Question: "who authored the taming of the shrew?"
Ground Truth: "william shakespeare"
Prediction: "william shakespeare"
Output: \{"score": 1, "explanation": "The prediction exactly matches the ground truth."\}

Question: "who authored The Taming of the Shrew?"
Ground Truth: "william shakespeare"
Prediction: "w shakespeare"
Output: \{"score": 1, "explanation": "The prediction matches the ground truth as w is the abbreviation of william."\}

Question: "what is the state bird of california?"
Ground Truth: "california quail"
Prediction: "california valley quail"
Output: \{"score": 1, "explanation": "The prediction matches the ground truth as california quail is also known as california valley quail."\}

Question: "what country is maharashtra metro rail corporation limited located in?"
Ground truth: "india"
Prediction: "indianapolis"
Output: \{"score": 0, "explanation": "The prediction doesn't match the ground truth answers."\}

Question: "what was wolves's record in their last game in eng-premier league?"
Ground Truth: "w"
Prediction: "wolves's record in their last game in the eng-premier League was a 2-1 win over fulham."
Output: \{"score": 1, "explanation": "The prediction correctly answers the result is a win."\}

Question: "how many seconds is 3 minutes 15 seconds?"
Ground Truth: "195 seconds"
Prediction: "3 minutes 15 seconds is 195 seconds."
Output: \{"score": 1, "explanation": "The prediction matches ground truth as it correctly states the number of seconds asked in the question."\}

Question: "on which days did am distribute dividends in the last year?"
Ground Truth: "2023-01-24, 2023-04-25, 2023-07-25, 2023-10-24"
Prediction: "the company distributed dividends on october 24, 2023 and january 24, 2023."
Output: \{"score": 0, "explanation": "The prediction partially matches the ground truth, but it is incomplete."\}

Question: "on which days did xxx distribute dividends in the last year?"
Ground Truth: "2023-01-13, 2023-03-25, 2023-11-21"
Prediction: "xxx distributed dividends on 1. 2023-01-13, 2. 2023-03-25, 3. 2023-10-21."
Output: \{"score": 0, "explanation": "The prediction contains an item, 2023-10-21, which is not in ground truth."\}

Question: "who are current band members of maroon 5?"
Ground Truth: "adam Levine, jesse carmichael, james valentine, matt flynn, pj morton, sam farrar"
Prediction: "they are jesse carmichael, matt flynn, adam levine, pj morton, sam farrar, and james valentine."
Output: \{"score": 1, "explanation": "The prediction exactly matches the ground truth."\}

Question: "which movies comprise the matrix franchise?"
Ground Truth: "the matrix, the matrix reloaded, the matrix revolutions, the matrix resurrections"
Prediction: "the matrix, the matrix reloaded, the matrix revolutions, the animatrix, and the matrix resurrections."
Output: \{"score": 0, "explanation": "The prediction covers more items than what are given by the ground truth."\}

Question: "how deep is the deepest lake of new york?"
Ground Truth: "618 ft"
Prediction: "the deepest lake in new york is seneca lake, with a depth of 618.23 feet."
Output: \{"score": 1, "explanation": "The prediction exactly matches the number in ground truth after rounding."\}

Question: "what is the closing price of meta yesterday?"
Ground Truth: "\$310.17"
Prediction: "310.2"
Output: \{"score": 1, "explanation": "The prediction exactly matches the number in ground truth after rounding."\}

Question: "what is the current market cap of appl?"
Ground Truth: "2.81 trillion"
Prediction: "2.667 trillion"
Output: \{"score": 0, "explanation": "The prediction does not match the number in ground truth."\}

Question: "what is the current pe ratio of appl?"
Ground Truth: "28.3"
Prediction: "the current pe ratio of apple is 26.66"
Output: \{"score": 0, "explanation": "The prediction does not match the number in ground truth."\}

Question: "how much is tesla's stock price down from its all-time high?"
Ground Truth: "\$221.83"
Prediction: "209.52"
Output: \{"score": 0, "explanation": "The prediction does not match the number in ground truth."\}

Question: "what is the length of amazon river?"
Ground Truth: "over 4000 miles"
Prediction: "the length of amazon river is 4,000 miles"
Output: \{"score": 0, "explanation": "The prediction does not say Amazon River is longer than 4000 miles."\}

Question: "how many copies x were sold?"
Ground Truth: "2 million."
Prediction: "it is over 2 million"
Output: \{"score": 0, "explanation": "The prediction does not match the ground truth."\}

Question: "what is the population of country x?"
Ground Truth: "3,576,873"
Prediction: "the population of country x is 3.3 million."
Output: \{"score": 0, "explanation": "The prediction, 3.3 M, does not match the number, 3.6 M, in ground truth."\}

Question: "what is the current market value of stock x?"
Ground Truth: "\$2,237,578,268"
Prediction: "\$2.16 billion."
Output: \{"score": 0, "explanation": "The prediction, 2.16 B, does not match the number, 2.2 B, in ground truth."\}

Question: "how old is michael jordan?"
Ground Truth: "61"
Prediction: "mj was born on February 17, 1963."
Output: \{"score": 0, "explanation": "The prediction does not explicitly say the age."\}

Question: "how long did amy teach as a teacher?"
Ground Truth: "24 years and 255 days"
Prediction: "25 years."
Output: \{"score": 0, "explanation": "The prediction does not exactly match the ground truth."\}

Question: "at what age did usher perform in the half time show of superbowl?"
Ground Truth: "45 years old"
Prediction: "usher has not performed at a super bowl halftime show."
Output: \{"score": 0, "explanation": "The prediction does not match the ground truth."\}

Question: "what year did olivia rodrigo win the grammys best new artist award?"
Ground Truth: "2022"
Prediction: "sure! here's an answer in three words or fewer: 2022 grammy awards."
Output: \{"score": 1, "explanation": "The prediction answers the correct number of the year."\}

Question: "how many times have zendaya and tom holland starred in a film together?"
Ground Truth: "3"
Prediction: "they have starred in 3 films together: spider-man: homecoming, spider-man: far from home, and spider-man: no way home."
Output: \{"score": 1, "explanation": "The prediction gives the right number and it is reasonable."\}

Question: "who is younger, timothee chalamet or tom holland?"
Ground Truth: "tom holland"
Prediction: "timothée chalamet is younger than tom holland."
Output: \{"score": 0, "explanation": "The prediction does not match the ground truth."\}

Question: "who had more number one hits on the us billboard, a or b?"
Ground Truth: "a had more number one hits on the us billboard than b, with 20 number one hits compared to b's 15."
Prediction: "a"
Output: \{"score": 1, "explanation": "The prediction is a concise and correct summary of the ground truth."\}

Question: "what is xxx's birthdate?"
Ground Truth: "1996-01-01."
Prediction: "02/01/1996"
Output: \{"score": 0, "explanation": "The prediction does not match the ground truth."\}

Question: "what was the worldwide box office haul for movie x?"
Ground Truth: "101756123."
Prediction: "102 million"
Output: \{"score": 1, "explanation": "The prediction exactly matches the number in ground truth after rounding."\}

Question: "how much has spotify's user base increased by since 2020 in na?"
Ground Truth: "spotify's user base increased by 34 million since 2020."
Prediction: "spotify's north american user base increased from 36 million in 2020 to 85 million by 2021"
Output: \{"score": 0, "explanation": "The prediction is not answering the question as it only gives the increase from 2020 to 2021."\}

Question: <\textit{Question}> 
Ground truth: <\textit{Ground Truth}>
Prediction: <\textit{Answer}>\\

\textit{\textbf{Auto-evaluation of Head2Tail}}\\
\noindent\rule{\linewidth}{0.4pt}

The task is provided a QUESTION with GROUND\_TRUTH answer, evaluate whether my ANSWER is correct, answer briefly with Yes/No. You will first see some [EXAMPLE]s on this task and then you will complete the [TASK].

Here are some examples:

[EXAMPLE]: QUESTION: Bangladesh Nationalist Party is the member of which international organization?
GROUND\_TRUTH: Asia Pacific Democrat Union or Centrist Democrat International
ANSWER: according to the wikipedia page, bangladesh nationalist party is a member of the centrist democrat international.
Your answer: Yes

[EXAMPLE]: QUESTION: What/who influenced Charles Fourier?
GROUND\_TRUTH: Nicolas-Edme Rétif
ANSWER: bob black.
Your answer: No

[EXAMPLE]: QUESTION: What patrol aircraft is used by the South African Air Force?
GROUND\_TRUTH: C-47 Skytrain
ANSWER: according to the wikipedia infobox, the patrol aircraft used by the south african air force is the c-47tp.
Your answer: Yes

[EXAMPLE]: QUESTION: Which automobile team had the fastest driver during the 1960 Indianapolis 500?
GROUND\_TRUTH: A.J. Watson
ANSWER: ken-paul
Your answer: No

[EXAMPLE]: QUESTION: What party was split from Communist Refoundation Party?
GROUND\_TRUTH: Italian Communist Party
ANSWER: according to the wikipedia infobox, the italian communist party (pci) was split from to form the communist refoundation party (prc) in 1991.
Your answer: Yes

[EXAMPLE]: QUESTION: Which company owns TV Land?
GROUND\_TRUTH: Paramount Media Networks
ANSWER: paramount global.
Your answer: No

[EXAMPLE]: QUESTION: What is the stadium where BSG Chemie Leipzig (1950)'s home matches are held?
GROUND\_TRUTH: Alfred-Kunze-Sportpark or Georg-Schwarz-Sportpark
ANSWER: alfred-kunze-sportpark (also known as georg-schwarz-sportpark)
Your answer: Yes

[EXAMPLE]: QUESTION: What language is spoken in Evening (magazine)?
GROUND\_TRUTH: Japanese language
ANSWER: english
Your answer: No

[EXAMPLE]: QUESTION: What is the ending theme of My Papa Pi?
GROUND\_TRUTH: Pia Wurtzbach
ANSWER: the ending theme of my papa pi is "my papa pi" by piolo pascual, pia wurtzbach, and pepe herrera.
Your answer: Yes

[EXAMPLE]: QUESTION: What is the record label for Cogumelo Records?
GROUND\_TRUTH: Relapse Records
ANSWER: cogumelo records.
Your answer: No

[EXAMPLE]: QUESTION: What is the legislative body in Albanian Kingdom (1928–1939)?
GROUND\_TRUTH: Parliament of Albania
ANSWER: according to the wikipedia infobox and summary, the legislative body in the albanian kingdom (1928–1939) is the constitutional assembly.
Your answer: Yes

[EXAMPLE]: QUESTION: Jim Pearson was born in which place?
GROUND\_TRUTH: Falkirk
ANSWER: chatham, ontario, canada.
Your answer: No

[EXAMPLE]: QUESTION: The predecessor of Cirilo Almario is?
GROUND\_TRUTH: Manuel del Rosario
ANSWER: manuel p. del rosario, d.d.
Your answer: Yes

[EXAMPLE]: QUESTION: What is the format of The Wedge (Australian TV series)?
GROUND\_TRUTH: Stereophonic sound
ANSWER: the format of the wedge (australian tv series) is a sketch show.
Your answer: No

[EXAMPLE]: QUESTION: What is the mouth of Montreal River (Timiskaming District)?
GROUND\_TRUTH: Timiskaming, Unorganized, West Part, Ontario
ANSWER: according to the wikipedia infobox and summary, the mouth of the montreal river (timiskaming district) is lake timiskaming on the ottawa river.
Your answer: Yes

[EXAMPLE]: QUESTION: Who developed Flappy?
GROUND\_TRUTH: DB-SOFT
ANSWER: flappy bird was developed by .gears, which is a game development company founded by dong nguyen.
Your answer: No

[EXAMPLE]: QUESTION: What significant design was created by Joseph Berlin?
GROUND\_TRUTH: Cinema of Israel
ANSWER: mograbi cinema, tel aviv.
Your answer: Yes

[EXAMPLE]: QUESTION: What is Cinematic soul derived from?
GROUND\_TRUTH: Disco
ANSWER: soul music, psychedelic soul, orchestral music, and film score.
Your answer: No

[EXAMPLE]: QUESTION: What patrol aircraft is used by the VPB-127?
GROUND\_TRUTH: Lockheed Ventura or PV-1
ANSWER: pv-1
Your answer: Yes

[EXAMPLE]: QUESTION: Which automobile team had the fastest driver during the 1953 Curtis Trophy?
GROUND\_TRUTH: Cooper Car Company
ANSWER: cooper-bristol.
Your answer: No

[TASK]: QUESTION: <\textit{Question}>
GROUND\_TRUTH: <\textit{Ground Truth}>
ANSWER: <\textit{Answer}>

Your answer is?

\end{block}

\end{document}